\documentclass[lettersize,journal,compsoc]{IEEEtran}
\usepackage{amsmath,amsfonts}
\usepackage{algorithmic}
\usepackage{algorithm}
\usepackage{array}
\usepackage{textcomp}
\usepackage{stfloats}
\usepackage{tikz}
\usepackage{url}
\usepackage{verbatim}
\usepackage{graphicx}
\usepackage{cite}
\usepackage{xcolor}
\usepackage{amsmath}
\usepackage{makecell}
\usepackage{multirow}
\usepackage{caption}
\usepackage{xspace}
\usepackage{subfigure}
\usepackage{ragged2e}
\usepackage{booktabs}

\newcommand{\etal}{\textit{et al.}}
\newcommand{\etc}{\textit{etc. }}

\newcommand{\method}{Sat2Density++\xspace}

\usepackage[colorlinks=true, linkcolor=blue, urlcolor=black]{hyperref}

\begin{document}
\title{Seeing through Satellite Images at Street Views}

\author{
Ming~Qian, \and 
Bin~Tan, \and 
Qiuyu~Wang, \and
Xianwei~Zheng, \and 
Hanjiang~Xiong, \and 
Gui-Song~Xia, \and
Yujun~Shen, \and
Nan~Xue

\IEEEcompsocitemizethanks{

\IEEEcompsocthanksitem M. Qian, X. Zheng, and  H. Xiong, are with the State Key Lab. LIESMARS, Wuhan University, Wuhan, 430079, China (e-mail: mingqian@whu.edu.cn; zhengxw@whu.edu.cn; xionghanjiang@whu.edu.cn).

\IEEEcompsocthanksitem B. Tan, Q. Wang,  Y. Shen, and N. Xue are with Ant Group, Hangzhou, 310013, China (e-mail: tanbin@whu.edu.cn; wangqiuyuu@gmail.com; shenyujun0302@gmail.com; xuenan@ieee.org).

\IEEEcompsocthanksitem   G.-S. Xia is with the School of Artificial Intelligence, Wuhan University, Wuhan, 430079, China (e-mail: guisong.xia@whu.edu.cn).

\IEEEcompsocthanksitem  Corresponding author: Xianwei Zheng and Gui-Song Xia.



}
}

\markboth{Accepted by IEEE TRANSACTIONS ON PATTERN ANALYSIS AND MACHINE INTELLIGENCE}%
{Shell \MakeLowercase{\textit{et al.}}: A Sample Article Using IEEEtran.cls for IEEE Journals}



\IEEEtitleabstractindextext{%
\begin{abstract}
\justifying 
This paper studies the task of SatStreet-view synthesis, which aims to render photorealistic street-view panorama images and videos given a satellite image and specified camera positions or trajectories. Our approach involves learning a satellite image conditioned neural radiance field from paired images captured from both satellite and street viewpoints, which comes to be a challenging learning problem due to the sparse-view nature and the extremely large viewpoint changes between satellite and street-view images. We tackle the challenges based on a task-specific observation that street-view specific elements, including the sky and illumination effects, are only visible in street-view panoramas, and present a novel approach, Sat2Density++, to accomplish the goal of photo-realistic street-view panorama rendering by modeling these street-view specific elements in neural networks. In the experiments, our method is evaluated on both urban and suburban scene datasets, demonstrating that Sat2Density++ is capable of rendering photorealistic street-view panoramas that are consistent across multiple views and faithful to the satellite image. Project page is available at \url{https://qianmingduowan.github.io/sat2density-pp/}.
\end{abstract}
}

\maketitle

\section{Introduction}\label{sec:intro}
\begin{figure*}[!ht]
    \includegraphics[width=1.0\linewidth]{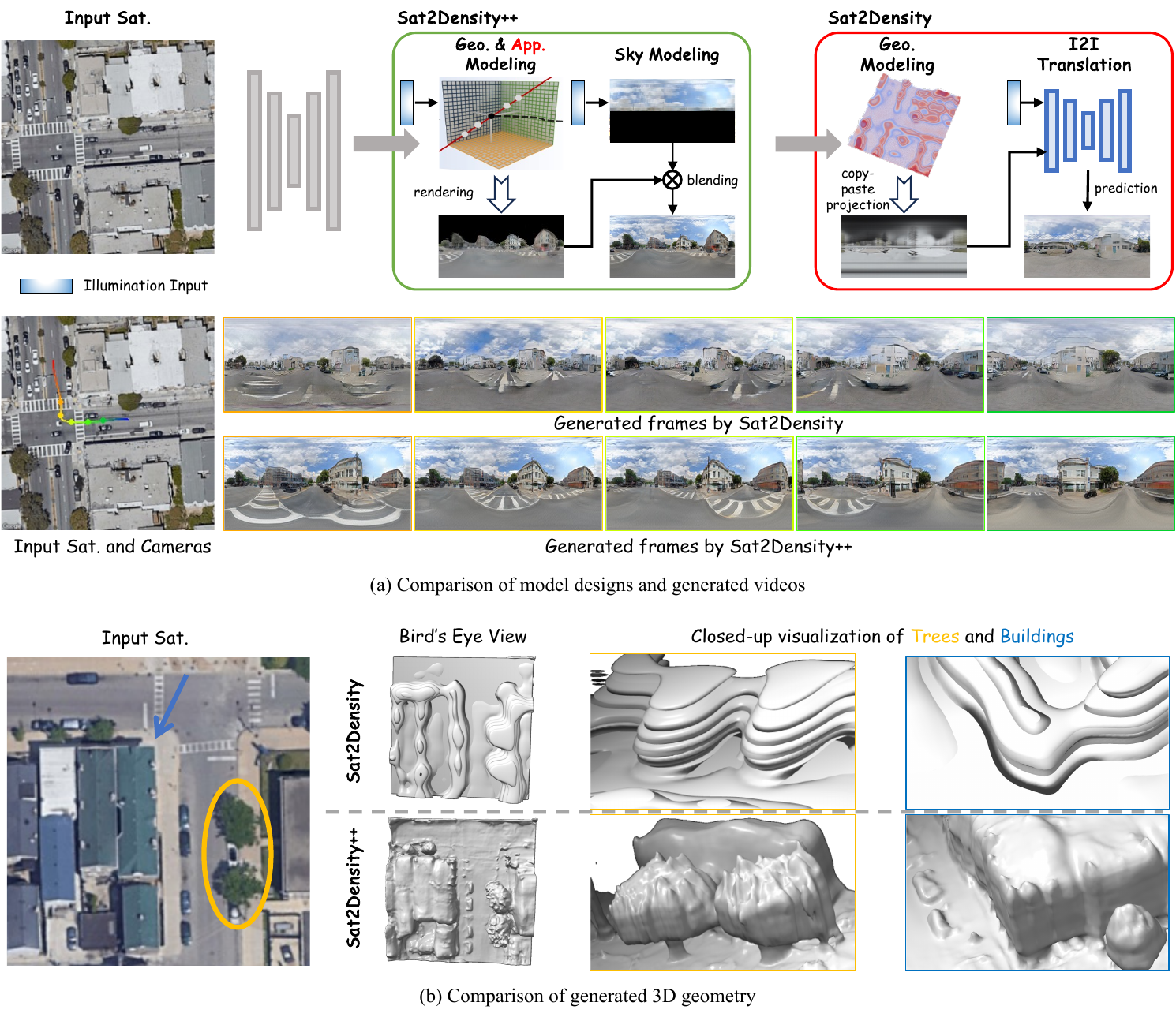}
    \vspace{-17pt}
    \captionsetup{type=figure}
    \caption{
\textbf{Comparison of model designs and results between Sat2Density \cite{Sat2Density} proposed in our conference version and \method (Ours) proposed in this paper.} (a) Overview illustration of model designs and the video results. At the bottom of Fig. 1(a), the colored camera trajectory is shown on the left satellite image, and the corresponding street-view images generated by models are shown on the right.
(b) A comparison of the generated geometry, where surfaces are extracted from the density field using Marching Cubes. 
{\em Please visit our \href{https://qianmingduowan.github.io/sat2density-pp//}{\texttt{project page}} for full video results.}
}
    \label{fig:teaser2}
\end{figure*}

\IEEEPARstart{R}econstructing 3D scenes from multiple views of images is a core problem in the 3D vision and photogrammetry communities. This problem has been extensively studied in neural rendering~\cite{NERF,gaussian,URF}, but usually suffers from a high demand of dense input views, often making it a resource-intensive practice, especially for outdoor scenes.
In this paper, we focus on the 3D reconstruction of outdoor scenes in neural rendering, but aim to reduce the required number of input views to just two or even one using the wide-coverage characteristics of satellite images.
We use a challenging two-view configuration that contains {\em one satellite image capturing a large-scale scene from a distant location and one street-view panorama capturing the same scene at street level}, with the goal of learning a feedforward model that captures faithful 3D geometry and renders high-quality street-view videos at any queried bird's-eye view~(BEV) trajectory for the given satellite image.

We refer to the two-view data as the \texttt{SatStreet}-view images, which pose an extremely challenging scenario for the establishment of visual correspondences and the neural rendering, 
but are viable by mutually and manually checking the context over the image pairs because both the satellite and street-view images are observing the same geographical contents.
Thus, we are curious about how we could infer the 3D scene geometry by curating the \texttt{SatStreet}-view image data.
We take the novel-view synthesis as the main goal and refer to the task as the \texttt{SatStreet}-view synthesis. 

The \texttt{SatStreet}-view synthesis task aims to render photorealistic street-view panorama images and videos given a satellite image and specified camera positions or trajectories, as shown in Fig.~\ref{fig:teaser2}.
\footnote{In this paper, all visualized satellite images are oriented with North at the top and West to the left. For the generated 360° panorama images, North aligns with the central axis, South spans the boundary edges, while West and East occupy the left and right quarter sections, respectively.} 
The primary challenge of this task stems from the significant viewpoint changes between satellite and street-level perspectives. 
Satellites capture scenes from a bird's-eye view, mainly focusing on the tops of buildings, road surfaces, trees, and other geographical features. 
In contrast, the panorama images provide street-level views from within the scene, offering side perspectives of buildings, detailed road surfaces, the intricate features of tree trunks and branches, \etc 
Additionally, street-view panorama images include elements such as the sky and the view-specific illumination effects, which are beyond the limited scope of satellite imagery. 
To address these substantial viewpoint changes, we focus on the 3D scene geometry, which is likely invariant to observational perspectives.

A robust 3D representation is essential for models to understand variations in camera viewpoints, thereby improving their ability to generate images more accurately.
Shi~\etal~\cite{Shi2022pami} proposed an end-to-end learning approach for height maps as an intermediate 3D geometric representation, demonstrating better results compared to image-to-image translation methods~\cite{Pix2Pix,Regmi_2018_CVPR} for single image generation. 
Other works~\cite{Lu_2020_CVPR, sat2vid, li2024sat2scene} have circumvented the challenges of learning 3D scenes from \texttt{SatStreet}-view image data by employing off-the-shelf semantic and metric depth maps of satellite images as supervision signals, thereby obtaining 2.5D building voxels to serve as conditions for street-view images generation.
All those studies highlighted the necessity of 3D shape representation for \texttt{SatStreet}-view synthesis, but they are limited by the accuracy and coherence of the used (or learned) 3D representations.

We approach the goal of \texttt{SatStreet}-view synthesis by learning the radiance field representation~\cite{NERF} with a feedforward neural network  
from the \texttt{SatStreet}-view image pairs.
Along this direction, two main challenges remained to be solved: 
\begin{enumerate}
        \item how to effectively learn the neural fields as the primary 3D representation from \texttt{SatStreet}-view image pairs;
        \item how to handle street-view-specific elements such as the sky, illumination effects, and detailed structures that are not visible to satellites. 
\end{enumerate}
Our main idea is based on a holistic viewpoint for these two key challenges using generative adversarial learning to 
learn the faithful 3D scene representation from the appearance of \texttt{SatStreet}-view image pairs.

\noindent\textbf{Preliminary Findings and the Limitations.~}
Our preliminary version presents Sat2Density~\cite{Sat2Density}, which takes the density field representation as the 3D representation of the scene geometry and learns the density field from the photometric characteristics of the street-view panoramas. Two key ingredients, the {\em non-sky opacity loss} and {\em illumination modeling} drive the learning process,
and we project color information from the satellite image onto the 3D geometry representation to render the initial street-view images by volume rendering.
Because of the copy-pasting scheme of color projection, the initial projected panoramas are often imprecise, and an image-to-image (I2I) translation module built with convolutional neural networks is used for refinement. 
While Sat2Density successfully learned plausible density fields in suburban scenes, its performance in complex urban environments was unsatisfactory, as evidenced by both the noisy density fields and the generated street-view videos (see Fig.~\ref{fig:teaser2}).
This shortcoming mainly arises from three factors: the limited expressiveness of a pure density field representation, the inaccuracies introduced during the volume rendering step, and the reliance on a 2D image-to-image refinement process that hinders effective density field learning. As a result, the method struggles to produce multi-view consistent street-view videos along a trajectory.

\noindent\textbf{Method Overview of \method.~}
        We extend the Sat2Density to \method in this paper.
        Compared to Sat2Density~\cite{Sat2Density} that only learned the density field while ignoring color in the common neural field to reduce the complexity of the 3D representation,
        in our \method, we demonstrate that both the scene geometry and the appearance modeling play key roles in modeling complex scenes, thus opt to learn an illumination-adaptive neural radiance field instead of the previously-used density field. This radiance field is capable of maintaining the characteristic hues of satellite
        imagery under neutral lighting conditions or adjusting the
        generated color features when explicit illumination information is available. 
        As a result, the density output of the illumination-adaptive neural radiance field is invariant across various lighting scenarios.  Building on the joint modeling of scene geometry and the appearance, we simplify the generation of street-view panoramas in two parallel branches: one for the ground scene rendering from the learned neural radiance field and the other for the sky region generation from the illumination-sensitive 2D sky region generation network, and then compose the final street-view panoramas from the output of the two branches. By joint training of all the neural network components with carefully-designed learning objectives, we are able to learn the faithful 3D scene geometry and generate high-quality street-view panoramas that are consistent across multiple views and faithful to the satellite image, and thus can consecutively render street-view videos for any queried BEV trajectory.
        Fig.~\ref{fig:teaser2} illustrates the case for the urban scene modeling that is challenging the existing approaches (including our preliminary Sat2Density~\cite{Sat2Density}) for \texttt{SatStreet}-view synthesis, in which our proposed \method learns faithful 3D geometry and renders high-fidelity street-view videos at any queried BEV trajectory.

To the best of our knowledge, our model is the first that can synthesize multi-view consistent street-view videos from input satellite images without relying on 3D annotations for training.
Through extensive qualitative and quantitative experiments, we demonstrate that \method outperforms existing single-image generation models in terms of image quality and shows enhanced geometric accuracy and significantly improved video faithfulness, consistency, and quality compared to our conference model. Besides, the predicted satellite view depth, the visualization of interpolated illumination, and the performance on out-of-distribution data further validate the reliability of \method.

\section{Related Works}

\subsection{Satellite-Ground Cross-view Perception}
Both street-view and satellite images provide unique perspectives of the world, and their combination provides us with a more comprehensive way to understand and perceive the world from satellite-ground visual data. However,  the drastic viewpoint changes between the satellite and ground images pose several challenges in geo-localization~\cite{CVUSA, shi2019spatial, shi2020looking, shi2020optimal, Unleashing_2024_CVPR,ji2025game4loc,song2023learning,xia2025fg,shi2022beyond,fervers2023uncertainty,sarlin2023orienternet,shi2023boosting,xia2023convolutional,shi2022cvlnet}, SatStreet-view synthesis\cite{Shi2022pami, sat2vid,Lu_2020_CVPR,tang2019multi,li2024sat2scene,li2024crossviewdiff,ze2025controllable}, cross-view synthesis~\cite{regmi_cross-view_2019, Regmi_2018_CVPR,lin2024geometry,ye2024skydiffusion}, street-view image synthesis from multiple satellite images~\cite{xu2024geospecific,gao2025skyeyes}, overhead image segmentation with the assistance of ground-level images~\cite{workman2022revisiting}, geo-enabled depth estimation~\cite{workman2021augmenting,Enhancing-Monocular-Height}, and predicting ground-level scene layout from aerial imagery~\cite{zhai2017predicting,OmniCity}.

To jointly extract more information from SatStreet-view images,
many previous works have proposed various approaches to model and learn the drastic viewpoint changes, including the use of homography transforms~\cite{regmi_cross-view_2019}, additional depth or semantic supervision~\cite{tang2019multi, Lu_2020_CVPR,sat2vid}, transformation matrices~\cite{CVUSA}, and geospatial attention~\cite{workman2022revisiting}, among others. Despite effectiveness, most approaches mainly focus on modeling the feature correlations at the image level, while overlooking the intrinsic correspondence between the two views in 3D space.
A seminal attempt by Shi~\etal~\cite{Shi2022pami} utilizes height multi-plane images as the geometry representation to overcome these challenges, but 
fails to get a good quality 3D representation from cross-view datasets. Our preliminary work~\cite{Sat2Density} builds on this effort, revealing crucial insights that significant viewpoint variations and varying imaging conditions in cross-view images are the key factors hindering the learning of scene representation from cross-view images, and proposes approaches to aid in learning the scene representation.

This paper builds upon our previous endeavors and introduces a novel framework. \method, trained on paired aerial and ground images, harnesses the latent 3D correspondences within the cross-view pair data to learn intricate 3D geometries and appearances, heralding a significant leap towards reliable view-consistent video generation. 



\subsection{Scene Generation from Conditions}
Recent advancements in 3D scene generation leverage diverse data sources to generate outdoor environments. Zhou~\etal\cite{infinite_nature_2020} and Li~\etal\cite{li2022_infinite_nature_zero} developed techniques to extend natural images into trajectories of novel views using depth warping. Xiangli~\etal~\cite{xiangli2023assetfield} uses a semantic bird's-eye view (BEV) map as a condition for 3D outdoor scene synthesis. However, 2D semantic BEV maps have limitations in representing the 3D space, such as the inability to adequately depict vertically stacked objects.

In the urban context, innovations such as UrbanGIRAFFE~\cite{UrbanGIRAFFE}, Streetscapes~\cite{deng2024streetscapes}, and InfiniCity~\cite{lin2023infinicity} emphasize semantic voxel-based conditioning for scene creation. These 3D semantic voxels can comprehensively represent 3D semantics and occupancy information of the scene, reducing the learning burden on the model. However, annotating 3D semantic voxels requires extensive manual labor, which increases exponentially with the desire for more refined categories.

Satellite images provide a valuable and abundant supplementary data source for outdoor scene reconstruction tasks. The extensive availability of satellite imagery offers a global perspective with rich geometric and topological details, enabling comprehensive coverage and consistent structural information. This aerial vantage point facilitates a clearer understanding of the relationships between urban elements and the overall scene layout. By incorporating large-scale, high-resolution satellite imagery into the training process, models can potentially reduce the reliance on intricate semantic annotations while maintaining a high level of detail and consistency in the generated 3D representations. The vast quantity of satellite images available makes them a natural and promising choice for augmenting existing datasets and improving the performance of outdoor scene reconstruction algorithms.

Our method capitalizes on the strengths of satellite imagery, utilizing it as a source of rich environmental context and a condition for precise 3D geometry estimation. By integrating satellite imagery with advanced machine learning techniques, our model aims to achieve a higher degree of quality in synthesizing street views that correspond seamlessly with their satellite counterparts.

\section{The Proposed \method}
\begin{figure*}[!t]
    \centering
    \includegraphics[width=0.9\textwidth]{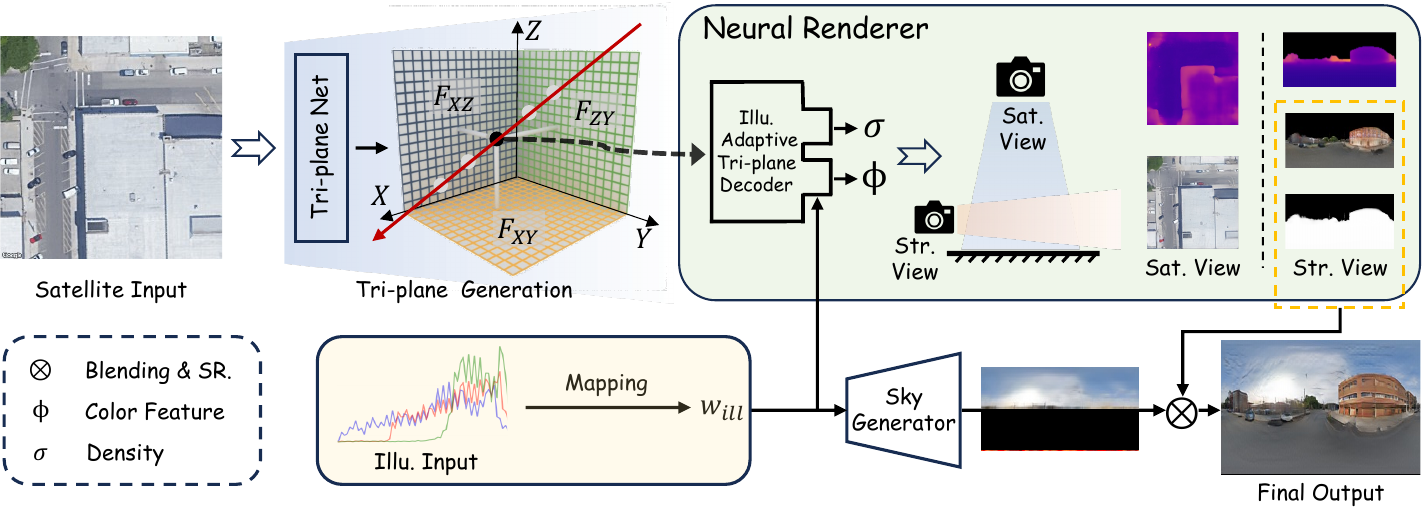}\\
    \vspace{-1mm}
    \caption{
    \textbf{Diagram of the proposed \method framework.} The system begins with a satellite image input, and it generates tri-plane features via the Tri-plane Net. Given specific camera poses, these features are then processed by an Illumination Adaptive Tri-plane Decoder within a Neural Renderer to render both the satellite image and the ground part of the street-view image. Additionally, a 2D sky generation module is responsible for creating the sky region in the street-view image. The final street-view images are obtained by first alpha-blending the ground and sky components, followed by super-resolution enhancement. The illumination input facilitates the rendering process by harmonizing both the Tri-plane Decoder and the 2D sky generation module. For clarity, we have omitted the steps involving the use of camera poses to generate image features from the radiance field and the super-resolution module, as well as the details of rendering from satellite viewpoints from the tri-plane.
    }
    \label{fig:Framework}
    \vspace{-1em}
\end{figure*}

This section introduces the details of our proposed \method,
as illustrated in Fig.~\ref{fig:Framework}. 
In our model, the roles of the satellite image and the sky illumination input are completely decoupled. The satellite image is responsible for generating a learnable implicit 3D representation of outdoor ground scenes. Using the given camera parameters for the satellite or panoramic street view, it can accurately produce the corresponding ground content. In contrast, the infinite sky region is treated as a 2D background panoramic image, which is generated using the sky illumination input. The sky illumination input simulates the effects of varying daylight conditions on photographs in outdoor environments, with different daylight scenarios altering both the sky appearance and impacting the imaging of ground regions. Consequently, in our model, the sky illumination input also influences the appearance of images when rendering the street region from the implicit 3D representation, allowing for dynamic adaptation to different lighting conditions.

\subsection{The 3D Representation}
In this section, we build the 3D representation for feedforward SatStreet-view synthesis. Our principle is that the satellite images determine the major 3D representation, and the sky region of street-view panoramas sets the view-specific illumination conditions. Both the 3D representation and illumination establish the foundation for accurate learning from SatStreet-view training data.

\subsubsection{Tri-plane Generation from Satellite Image}
Given a satellite image $I_\text{sat} \in \mathbb{R}^{3\times256\times256}$ as the input, our \method leverages a neural network Tri-plane Net $G_{tri}$, to extract the image features $F_{\text{img}} \in \mathbb{R}^{96\times256\times256}$ by 
\begin{equation}
F_{\text{img}}=G_{\rm tri}(I_\text{sat}).
\end{equation}

Then, $F_{\text{img}}$ is split and reshaped to form the tri-plane feature maps $F_{tri}=\{ F_{XY}, F_{ZY}, F_{XZ} \}$ by 
\begin{equation}
    F_{tri}=\{ F_{XY}, F_{ZY}, F_{XZ}\} = {\rm Split}(F_{\rm img}),
\end{equation}
in which the size of each map is $32\times256\times256$. 
The tri-plane feature maps are utilized to represent the 3D scene within a tri-plane-based radiance field model~\cite{eg3d2022}.


Following the principles of the radiance field~\cite{max1995optical,NERF,eg3d2022}, given any street-view camera poses, we use bilinear interpolation to query point features \( f_{\text{tri}} \in \mathbb{R}^{96} \) from \( F_{\text{tri}} \) for the sampled 3D location along each camera ray. These point features are then sent to the illumination-adaptive tri-plane decoder (Sec.~\ref{Sec.Tri-Plane Decoder}) to obtain the density and color values for each point, which are subsequently used to render the ground part of street-view images.



\subsubsection{Sky Illumination Modeling from Street-View Image}

As our goal is to learn how to synthesize street-view panoramas from satellite images, it is essential to model illumination conditions that are specific to street-level observations. 
Illumination is closely related to the sky region in street-view images \( I_{\text{str}} \). Therefore, we adopt a statistical approach to model the histogram-based illumination feature from each street-view image, which efficiently encodes color information while maintaining computational simplicity, interpretability, and does not rely on a deep learning process. 
Lalonde et al. \cite{lalonde2007photo} also utilized sky histograms to match global lighting conditions.

In detail,
for each $I_{\rm str}$, we first extract binary sky masks $M_{\rm sky}$ using an off-the-shelf segmentation model~\cite{zhang2022bending}. We then featurize the sky pixels into color distributions to capture illumination characteristics. For this featurization, we adopt a histogram-based representation.
During training, the extraction of color histograms from sky pixels is straightforward. We compute separate histograms for the R, G, and B channels using 90 bins uniformly spanning the range $[0, 255]$. Each histogram is normalized by the total number of sky pixels, and the three normalized histograms $\tilde{H}_R$, $\tilde{H}_G$, and $\tilde{H}_B$ are concatenated in R-G-B order to form the illumination feature $f_{\rm ill}$. Formally, this process is expressed as:
\begin{equation}
f_{\rm ill} = {\rm concat}(\tilde{H}_R,\tilde{H}_G,\tilde{H}_B) \in \mathbb{R}^{270},
\end{equation}
where $\tilde{H}_R, \tilde{H}_G, \tilde{H}_B$ denote the normalized histograms of the red, green, and blue channels, respectively. In cases where no sky region is detected in the panorama, we set $f_{\rm ill} = \mathbf{0}$.

\noindent\textbf{Remarks on the Inference Process.}
Leveraging our approach, which decouples the sky illumination module from 3D implicit scene generation, we are able to produce street-view images or videos under various daylight conditions from a single satellite image by applying different sky illumination inputs. In our quantitative tests, we adopted multiple strategies. For instance, results marked with `*’ in Tab.~\ref{tab:result} illustrate that each satellite image was paired with a randomly selected sky illumination from the training set for rendering street views. Conversely, results marked with `{\dag}’ in Tab.~\ref{tab:result} utilized sky illumination inputs derived from the corresponding ground truth street-view images. In our qualitative results, we further showcased the capability to generate street views under different lighting conditions while maintaining a constant satellite image, as depicted in Fig.~\ref{fig:single compare} and Fig.~\ref{fig:illumination}.

\subsubsection{Illumination-Adaptive Tri-Plane Decoder
}\label{Sec.Tri-Plane Decoder}

Given the tri-plane feature $f_{\rm tri}$ for each sampled 3D location, we set up an illumination-adaptive tri-plane decoder for the computation of density value $\sigma\in\mathbb{R}$ and the appearance feature $\phi\in\mathbb{R}^{32}$ by 
\begin{equation}\label{eqn: mlp}
    \phi, \sigma = {\rm Decoder}(f_{\rm tri}, \mathbf{w}_{\rm ill}),
\end{equation}
where $\mathbf{w}_{\rm ill}\in\mathbb{R}^{512}$ is a style vector transformed from the illumination feature $f_{\rm ill}$ by an illumination mapping layer $E_{\rm ill}(\cdot)$ that is implemented by $8\times$ fully connected layers, read as 
\begin{equation}
    \mathbf{w}_{\rm ill} = E_{\rm ill}(f_{\rm ill}).
\end{equation}

The vanilla tri-plane decoder consists of a two-layer structure formed by multilayer perceptrons (MLPs). In this model, $f_{\rm tri}$ are processed through these MLPs, producing $\phi$ and $\sigma$ outputs simultaneously in the final layer. 
This approach assumes consistent lighting conditions across all viewpoints, providing a straightforward output based solely on triplane features~\cite{eg3d2022}.

In contrast, the illumination adaptive tri-plane decoder modifies this structure to address varying lighting conditions more effectively. Instead of using two stacked MLP layers as in the vanilla version, the illumination adaptive version replaces the final layer with two distinct branches for generating density and color, respectively. The density branch utilizes an MLP that takes the output from the first MLP layer as input to produce $\sigma$. The color branch also uses an MLP, but it concatenates the output from the first MLP layer with the illumination vector $\mathbf{w}_{\rm ill}$ before processing to generate $\phi$. This design allows the color features to be adaptively adjusted according to lighting conditions, offering more nuanced control and improving the accuracy and realism of visual representation under different illumination contexts.

\subsection{Street-View Image Generation}\label{sec.Neural Renderer}

With the illumination-adaptive tri-plane representation, we focus on the generation of street-view panoramas by first leveraging volumetric rendering to yield the ground part and then generating a sky image. An alpha blending of the rendered ground part and the generated sky part produces the final street-view panoramas. 
\subsubsection{Ground Part Generation}

\noindent\textbf{Camera Definition.}
The panorama street-view camera utilizes a cylindrical projection model, positioning it horizontally with the camera’s compass heading aligned to true north. In the absence of height data for the street-view camera positions, we assume that all street images are captured at a uniform height above the ground.

\noindent\textbf{Volume Rendering.}
Given a camera \( P \), we sample \( N \) points along each ray \( r \) through each pixel that emanates from the camera location. Let \( x_i \) be the \( i \)-th sampled point along the ray \( r \), we first compute density value \( \sigma_i \), and color feature \( \phi_i \) with the illumination-adaptive tri-plane decoder. Then, we use volume rendering\cite{max1995optical,NERF} to calculate the image feature map \( \hat{I}_F \), opacity map \( \hat{O} \), and depth map \( \hat{D}  \) along ray $r$ as follows:
\begin{equation}
\hat{I}_F(r) = \sum_{i=1}^N \tau_i \phi_i,
\end{equation}
\begin{equation}
\hat{O}(r) = \sum_{i=1}^N \tau_i,
\end{equation}
\begin{equation}
\hat{D}(r) = \sum_{i=1}^N \tau_i d_i,
\end{equation}
where \( d_i \) is the distance between the camera position and the point position \( i \), and the transmittance \( \tau_i \) along the ray \( r \) is computed as the probability of a photon traversing between the camera center and the i-th point given the length of i-th interval $\delta_i$
\begin{equation}
\tau_i =  \prod_{j=1}^{i} \exp(-\sigma_j \delta_j)  \left( 1 - \exp(-\sigma_i \delta_i) \right),
\end{equation}
note that, the synthesized raw color $\hat{I}_C(r)$ is taken from the first three channels of $\hat{I_F}(r)$,  the other channels are used for the super-resolution (SR) module.

Finally,  given a street-view camera parameter, the model can generate ground part street-image feature $\hat{I}_{F\text{grd}} \in \mathbb{R}^{32\times H\times W}$,  image color $\hat{I}_{C\text{grd}} \in \mathbb{R}^{3\times H\times W}$, opacity $\hat{O}_{\text{grd}} \in \mathbb{R}^{1\times H\times W}$, and depth $\hat{D}_{\text{grd}} \in \mathbb{R}^{1\times H\times W}$, the rendered size $(H, W)$ depends on the given camera intrinsics, in practice, the rendered size is $(64,256)$.

\subsubsection{Sky Part Generation and Alpha Blending}\label{sec.sky generator2}
We generate the sky part street-view image feature  $\hat{I}_{F{\text{sky}}}\in \mathbb{R}^{32\times H\times W}$ from the illumination vector $\mathbf{w}_\text{ill}$ by a 2D sky generator $G_{\text{sky}}(\cdot)$ in
\begin{equation}
\hat{I}_{F\text{sky}}  = G_{\text{sky}}(\mathbf{w}_\text{ill}).
\end{equation}
In our implementation, the architecture of $G_\text{sky}$ follows the generator used in StyleGAN-2~\cite{stylegan2}.

After that, we compute the low-resolution street-view image feature $\hat{I}_{F\text{str}}$ by  alpha blending:
\begin{equation}
\hat{I}_{F\text{str}} = \hat{O}_{\text{grd}} \times \hat{I}_{F\text{grd}} + (1-\hat{O}_{\text{grd}}) \times \hat{I}_{F\text{sky}}.
\end{equation}

Following the strategy used in volume rendering, the synthesized blended street-view raw resolution image color $\hat{I}_{C\text{str}}$ is the first three layers of $\hat{I}_{F\text{str}}$. 




\noindent\textbf{Street-View Super Resolution.}
Similar to prior works~\cite{eg3d2022,kangle2023pix2pix3d,gu2021stylenerf,pan2021shadegan}, as a final step, we apply a 2D image super-resolution (SR) module  $U$ to reduce the computational cost for volume rendering when synthesizing street-view images. We obtain the high-resolution image $\hat{I}_{C\text{str}}^+$ by feeding low-resolution $\hat{I}_{F\text{str}}$ into a  lightweight SR module $U$:
\begin{equation}
\hat{I}_{C\text{str}}^+ = U(\hat{I}_{F\text{str}}).
\end{equation}

In summary, our method finally synthesizes street-view images in high resolution $\hat{I}_{C\text{str}}^+  \in \mathbb{R}^{3\times128\times512}$ by a camera parameter $P$.

\subsection{Training-Time Satellite-View Generation}

To enhance the appearance alignment between the learned 3D representation and input \( I_{\text{sat}} \), we render the satellite view during training to provide additional supervision.
Since existing datasets~\cite{CVUSA,Shi2022pami,zhu2021vigor} did not provide satellite camera intrinsics and extrinsics, we approximate the satellite camera as an orthographic camera oriented vertically downward toward the ground. With the camera approximation, we can compute the ray origins and ray directions corresponding to each pixel for the satellite view rendering. 

Unlike street-view image generation, the sky generation module and SR module are not required when generating satellite-view images. 
To deal with the absence of street-view grounded illumination, when synthesizing satellite-view images, we substitute the illumination style vector $\mathbf{w}_{\text{ill}}$ in Eq.~(\ref{eqn: mlp}) with a null-style vector $\mathbf{w}_0 = \mathbf{0}$, as the tri-plane feature map $F_{\text{tri}}$ is directly derived from input satellite imagery. This null-style vector serves as a fixed placeholder to disable unnecessary style modulation in the generation pipeline.
Finally, given satellite camera parameters, the model can generate satellite-view image feature $\hat{I}_{F\text{sat}}$, image color $\hat{I}_{C\text{sat}}$,  depth $\hat{D}_{\text{sat}}$ by volume rendering.

\subsection{Learning Objectives}\label{sec:loss1}
The training data consists of paired satellite and street-view images. We also provide extra binary pseudo sky masks $M_{\text{sky}}$ for each street-view image by an off-the-shelf sky segmentation model~\cite{zhang2022bending}. 
In this section, we introduce the learning objectives, including reconstruction, opacity, and discriminator losses. 
\subsubsection{Non-sky Opacity Loss}\label{sec:Pseudo Opacity Loss}
In \method, we aim to learn the 3D representation of the satellite scene, but street-level scenes introduce elements such as the sky, which are not accounted for in satellite images. This challenges the model’s ability to recognize and accurately model the geometric structure of scenes captured in satellite imagery. 
We found that the pseudo sky masks provide a strong inductive basis to regularize the density field in a simple way, thus using a binary cross-entropy (BCE) loss for the supervision:

\begin{equation}
\mathcal{L}_{\text{opa}} = 
\text{BCE}(\hat{O}_{\text{grd}},1- M_\text{sky}),
\end{equation}
where $M_{\text{sky}}$ is the ground-truth sky mask,  1 indicates the sky region, and 0 indicates the ground region.

\subsubsection{Reconstruction Loss}\label{sec:Reconstruction Loss}
We introduce three reconstruction loss terms to supervise the quality of the generated satellite-view image $\hat{I}_{C\text{sat}}$, sky image $\hat{I}_{C\text{sky}}$, and whole street-view image $\hat{I}_{C\text{str}}^+$.

For the generated sky image $\hat{I}_{C\text{sky}}$, there is no available ground-truth sky image for supervision.
We calculate the \( \ell_1 \) loss between the predicted sky image \( \hat{I}_{C\text{sky}} \) and the real street image \( I_\text{str} \) exclusively at the pixel locations specified by the sky mask \( M_{\text{sky}} \). This focus ensures that the loss computation targets only the sky regions, allowing the model to prioritize accurate sky reconstruction without being influenced by other parts of the image. The loss function is defined as follows:
\begin{equation} 
\mathcal{L}_{\text{sky}} = \left\| M_{\text{sky}} \odot \left( \hat{I}_{C\text{sky}} - I_\text{str} \right) \right\|_1,
\end{equation}
where the operator \( \odot \) denotes element-wise multiplication.

For the street-view rendering, we  combine use the perceptual LPIPS~\cite{LPIPS} loss and  $\mathcal{L}_{1}$ loss to minimize the reconstruction error between the generated high-resolution panorama $\hat{I}^+_{C{\text{str}}}$ and $I_{\text{str}}$ by
\begin{equation}
\mathcal{L}_{\text{str}} = 
\mathcal{L}_{1}(\hat{I}^+_{C{\text{str}}},I_\text{str}) +
\mathcal{L}_{\text{lpips}}(\hat{I}^+_{C{\text{str}}},I_\text{str}).
\end{equation}


Lastly,  we apply a satellite view reconstruction loss to align the generated 3D appearance with the input satellite image $I_{sat}$. The satellite view reconstruction loss \( \mathcal{L}_{\text{sat}} \) is computed by:
\begin{equation}
\mathcal{L}_{\text{sat}} = \mathcal{L}_{1}(\hat{I}_{C\text{sat}},I_{sat}) + \mathcal{L}_{\text{lpips}}(\hat{I}_{C\text{sat}},I_{sat}).
\end{equation}

The final reconstruction loss is weighted by
\begin{equation}
\mathcal{L}_{\text{recon}} =  \lambda_{\text{sat}}\mathcal{L}_{\text{sat}} +
\lambda_{\text{str}}\mathcal{L}_{\text{str}} + \lambda_{\text{sky}}\mathcal{L}_{\text{sky}},
\end{equation}
where $\lambda_{\text{sat}}$, $\lambda_{\text{str}}$, and $\lambda_{\text{sky}}$ balance three terms. Besides, to save consumption, when synthesizing satellite view images, we render at one-quarter resolution with random crops during training.

\subsubsection{Discriminator Loss}\label{sec:Discriminator}
The reconstruction loss alone fails to synthesize detailed results from novel viewpoints. Therefore, we use an adversarial loss~\cite{GAN} to enforce renderings to look realistic from rendered viewpoints.
Specifically, we have two discriminators $D_{\rm str}$ and $D_{sat}$ for street-view images and satellite-view images, respectively.

The $D_{\rm str}$ follows the dual-discrimination design in  EG3D~\cite{eg3d2022}, discriminates on a 6-channel concatenation of the final image and the raw neural rendering, to maintain consistency between high-resolution final images and view-consistent (but low-resolution) neural renderings.
The total adversarial loss is written as follows,
\begin{equation}
\begin{aligned}
\mathcal{L}_{\text{GAN}} &= \lambda_{\rm 
 D_{\text{str}}}\mathcal{L}_{\rm 
 D_{\text{str}}}(\hat{I}_{C\text{str}}^+,\hat{I}_{C\text{str}}) 
+ \lambda_{\rm 
 D_{\text{sat}}}\mathcal{L}_{\rm 
 D_{\text{sat}}} (\hat{I}_{C\text{sat}}),
\end{aligned}
\end{equation}
where $\lambda_{\rm 
 D_{\text{str}}}$ and $\lambda_{\rm 
 D_{\text{sat}}}$ are hyperparameters for loss weighting, $\hat{I}_{C\text{str}}$ is the first three layers of blended street-view panorama raw feature $\hat{I}_{F\text{str}}$.
The non-saturating GAN loss function~\cite{GAN} with R1 regularization~\cite{DBLP:conf/icml/MeschederGN18} is used for $\mathcal{L}_{\text{GAN}}$, following the training scheme in StyleGAN2~\cite{stylegan2}.

\subsubsection{Total Loss}
Our final learning objective is written as follows:
\begin{equation}
\mathcal{L}_{\text{total}} = \mathcal{L}_{\text{recon}} + \lambda_\text{opa}\mathcal{L}_{\text{opa}}  + \mathcal{L}_{\text{GAN}}.
\end{equation}

\begin{figure*}[ht]
    \centering
    \includegraphics[width=0.95\textwidth]{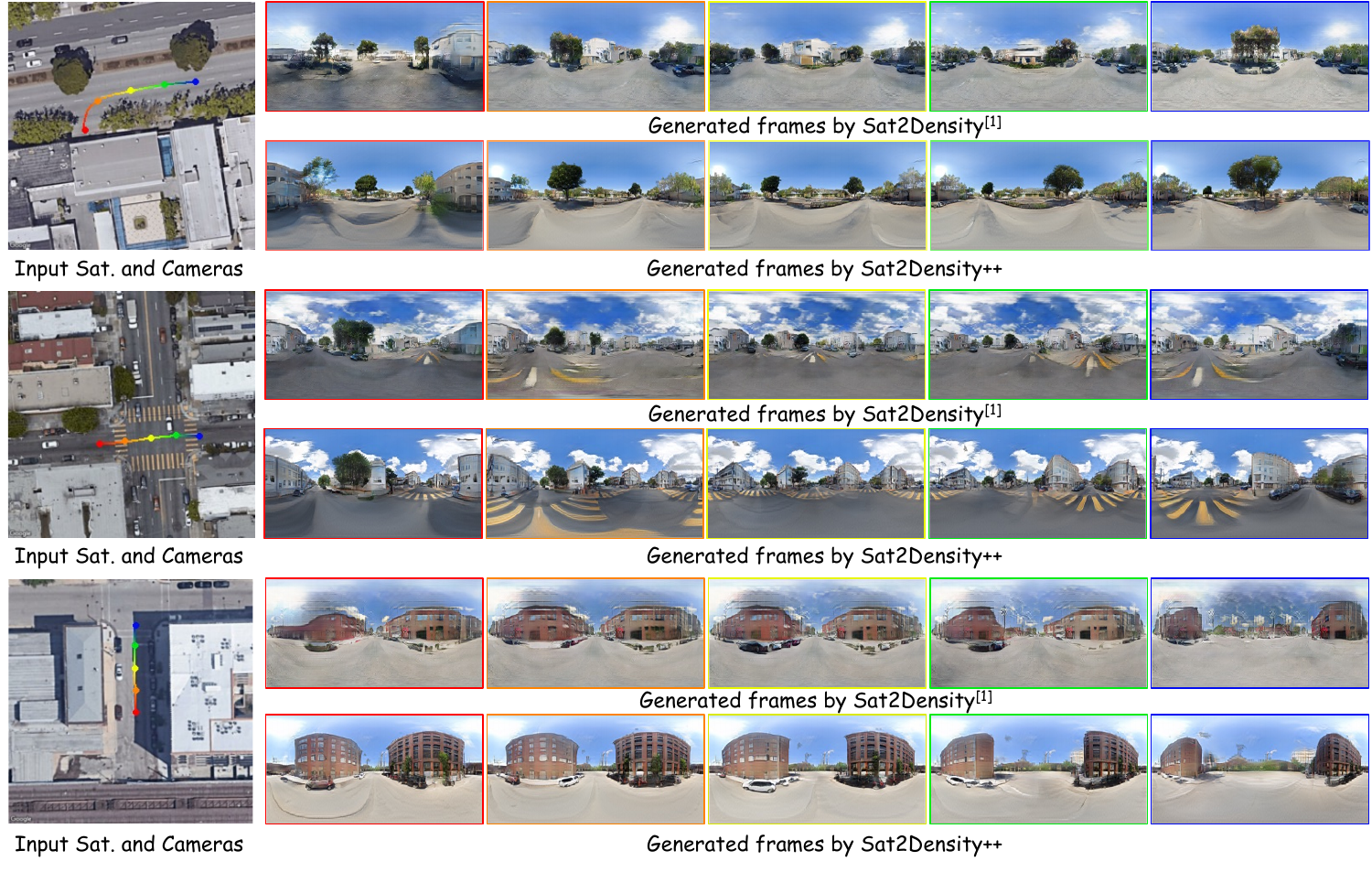}\\
    \vspace{-1mm}
    \caption{ 
    Three video results generated by our method and Sat2Density\cite{Sat2Density} on the VIGOR dataset. 
    For each example, the left panel shows the satellite image with the camera trajectory, which goes from the red point to the blue point. 
    Along this trajectory, we select five camera positions (highlighted points) to illustrate multi-view continuity. 
    The five groups of panoramas on the right correspond to these five camera positions in order. 
    In each group, the upper row shows the results of Sat2Density, while the lower row shows the corresponding results of our method at the \emph{same camera positions}, ensuring a fair one-to-one comparison. {\em The full videos can be seen on the \href{https://qianmingduowan.github.io/sat2density-pp//}{\texttt{project page}}.}
    }
    \label{fig:demo_vigor}
    \vspace{-1em}
\end{figure*}

\begin{figure*}[ht]
    \centering
    \includegraphics[width=0.95\textwidth]{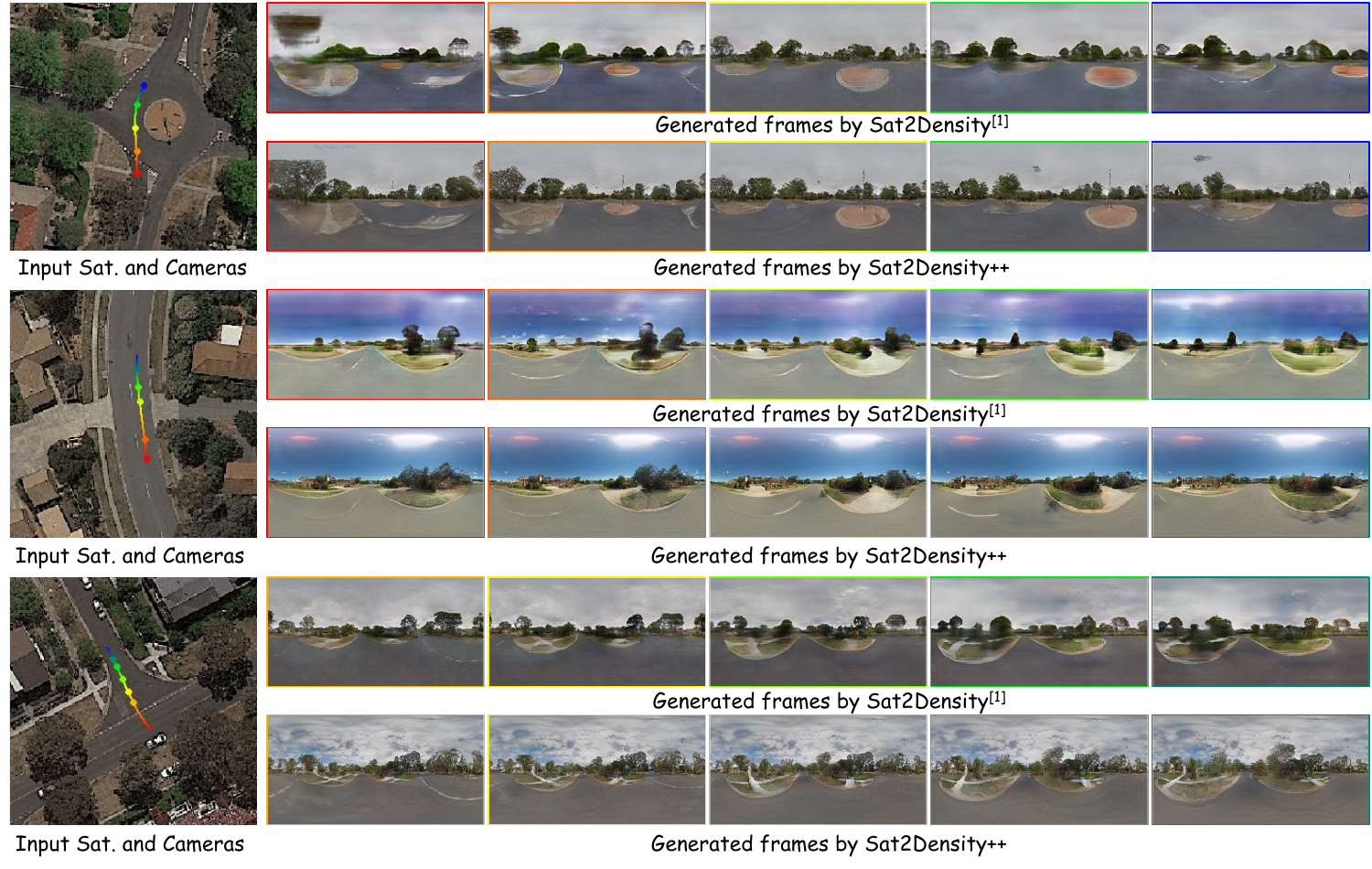}\\
    \vspace{-1mm}
    \caption{
    Three video results generated by our method and Sat2Density~\cite{Sat2Density} on the CVACT dataset. 
    For each example, the left panel shows the satellite image with the camera trajectory, which goes from the red point to the blue point. 
    Along this trajectory, we select five camera positions (highlighted points) to demonstrate the multi-view continuity around roundabouts or road intersections. 
    The five groups of panoramas on the right correspond to these five camera positions in order. 
    In each group, the upper row shows the results of Sat2Density, while the lower row shows the corresponding results of our method at the \emph{same camera positions}, ensuring a fair one-to-one comparison.
    {\em The full videos, including all sampled positions, can be found on the \href{https://qianmingduowan.github.io/sat2density-pp//}{\texttt{project page}}.}}    
    \label{fig:demo}
\end{figure*}

\begin{figure}[ht]
    \centering
    \includegraphics[width=0.45\textwidth]{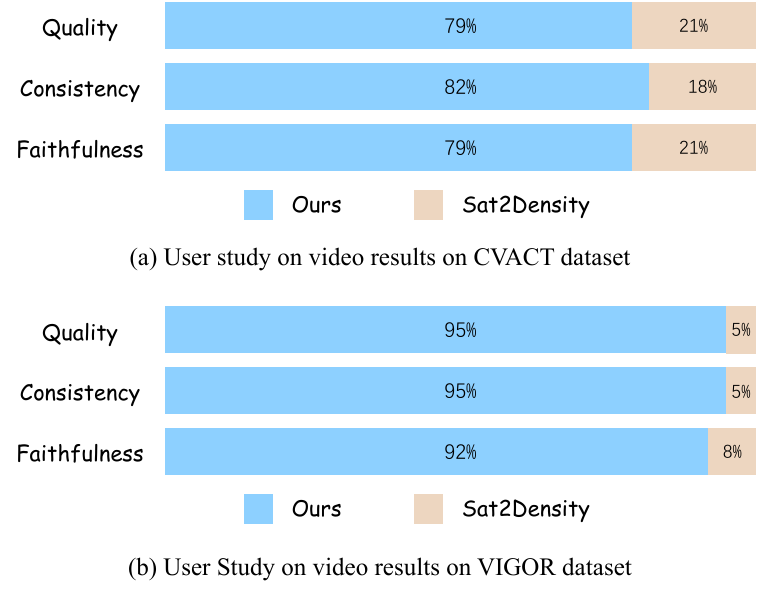}\\
    \vspace{-1mm}
    \caption{\textbf{Comparison of Ours and Sat2Density in User Studies on Video Results.} Users evaluated the Quality, Consistency, and Faithfulness of the generated videos by observing the input satellite images paired with the corresponding camera trajectory videos, as well as the videos produced by Sat2Density and \method. They compared the methods based on these three criteria and selected the results they found to be superior. Finally, we aggregated the average preferences across multiple video sets to determine the overall user preference levels. }
    \label{fig:user_study}
    \vspace{-1em}
\end{figure}

\section{Experiments}

\subsection{Datasets}\label{sec.dataset}

 In our experiments, we comprehensively evaluate our method on both suburban and urban scenes using three datasets: CVUSA~\cite{CVUSA}, CVACT~\cite{Shi2022pami}, and VIGOR~\cite{zhu2021vigor}.

\noindent\textbf{CVACT~\cite{Shi2022pami} and CVUSA~\cite{CVUSA} for Suburban Scenes.}
 The CVACT and CVUSA datasets consist of one-to-one paired satellite and street-view images. Each street-view image is assumed to be captured from the center of its corresponding satellite image. Following the recommended data splits from \cite{Shi2022pami}, the CVACT dataset includes 26,519 training samples and 6,288 testing samples, whereas the CVUSA dataset includes 35,532 training samples and 8,884 testing samples. 
 
\noindent\textbf{VIGOR for Urban Scenes.}
 The VIGOR dataset significantly differs from CVACT and CVUSA in several key aspects. First, VIGOR comprises urban scenes collected from major cities, including Chicago, New York City, San Francisco, and Seattle. Compared to suburban datasets, the complex structures found in urban environments pose greater challenges to existing methods~\cite{Sat2Density}. Second, each VIGOR data sample consists of one satellite image paired with two to three street-view panoramas, introducing an additional challenge related to viewpoint variations among multiple street-view captures per satellite location. Our method, through precise learning of 3D geometry, effectively addresses both structural complexities and viewpoint variations simultaneously. 
 
 We preprocess the data samples in VIGOR dataset to SatStreet image pairs for both training and testing. 
 We train our method on 40,733 pairs from Chicago, New York City, and San Francisco. To comprehensively evaluate performance, we establish two protocols: in-domain testing and zero-shot evaluation. For in-domain testing, we utilize 5,000 image pairs from the same cities to assess our model's effectiveness. Additionally, we perform zero-shot evaluation using 11,875 pairs from Seattle to examine our method's generalization capability in unseen environments. For accurate street-view image localization, we utilize revised GPS annotations from SliceMatch~\cite{slicematch} during data preprocessing.

\subsection{Implementation Details}\label{sec.model detail}
Our code is built on Carver~\cite{Carver}, an efficient PyTorch-based library for training 3D-aware generative models. 
For hyperparameters, we empirically assign the values of $\lambda_{\rm 
 D_{\text{str}}}$, $\lambda_{\rm 
 D_{\text{sat}}}$, $\lambda_\text{sat}$, $\lambda_\text{str}$, $\lambda_\text{sky}$, and $\lambda_\text{opa}$ to 1.0, 1.0, 30.0, 10.0, 10.0, and 25.0, respectively. 
We use 8 NVIDIA A100 GPUs for training, with a batch size of 4 per GPU. 
For the CVACT and CVUSA datasets, the training involves 180,000 iterations and takes about 30 hours. In contrast, training on the VIGOR dataset involves 360,000 iterations over an extended duration of approximately 60 hours.

\subsection{Evaluation Protocols and Metrics}\label{sec.metric}
The evaluation metrics used in our experiments comprehensively measure the quality of the generated street-view images and videos, which include FID and KID for realism evaluation and DINOv2 similarity for the high-level consistency, as well as the conventionally used SSIM~\cite{SSIM}, PSNR~\cite{PSNR}, and LPIPS~\cite{LPIPS}. For the evaluation of generated videos, due to the absence of video ground truth, user studies on the CVACT and VIGOR datasets are used to measure the quality, consistency, and faithfulness of videos.

\noindent\textbf{FID and KID.} Fr\'{e}chet Inception Distance (FID)~\cite{FID} and Kernel Inception Distance (KID)~\cite{KID} measure the distribution differences between the generated images and the ground truth. Other than the per-image evaluation, the comparisons on distribution differences reflect the overall plausibility of the generated samples. In our experiments, we use these two metrics for realism evaluation, and they are computed using the Clean-FID library~\cite{parmar2021cleanfid}.

\noindent\textbf{DINOv2 Similarity.} Pretrained DINOv2 models~\cite{oquab2023dinov2} have been identified as effective high-level feature representations of images. We use the pretrained DINOv2-L model to extract the $N$ visual tokens $\{\mathbf{f}_{1,i}\}_{i=1}^N$ from a generated image and $\{\mathbf{f}_{2,i}\}_{i=1}^N$ from its ground truth, then compute their average cosine similarities by 
\begin{equation}
    \frac{1}{N} \sum_{i=1}^N \frac{\mathbf{f}_{1,i}^T\mathbf{f}_{2,i}}{\|\mathbf{f}_{1,i}\|\|\mathbf{f}_{2,i}\|}
\end{equation}
as the overall similarity of them. 

\noindent\textbf{SSIM, PSNR, and LPIPS.} These three metrics follow the previous works~\cite{Shi2022pami,Sat2Density} to measure the quality of generated images. SSIM assesses the structural similarity, PSNR gauges the pixel-wise distance, and LPIPS~\cite{LPIPS} computes the high-level perceptual similarity between two images.  For the computation of LPIPS, we utilize the pre-trained AlexNet~\cite{AlexNet} and SqueezeNet~\cite{SqueezeNet} as the feature backbones, referred to as $P_{\text{alex}}$ and $P_{\text{squ}}$, respectively.

\noindent\textbf{User Studies for Generated Video.} Existing datasets consist solely of image collections and lack video data, so we quantitatively compare video results through a user study involving 31 participants and comprising 40 sets of videos. Each set is based on an input satellite image and camera trajectory video and includes videos generated by both our method and Sat2Density. 20 video sets are derived from the CVACT dataset, while the remaining are from the VIGOR dataset. Specifically, we assess the faithfulness of the generated videos to the input satellite images (Faithfulness), the consistency of the videos across multiple views (Consistency), and the overall visual quality exhibited by the generated content (Quality).

\subsection{Main Comparisons}\label{sec.comparision}
We compare \method with both image generation and video generation methods. 
For image generation methods, we evaluate against Pix2Pix~\cite{Pix2Pix}, XFork~\cite{regmi_cross-view_2019}, and Shi et al.~\cite{Shi2022pami}. For video generation methods, we compare with our preliminary work Sat2Density~\cite{Sat2Density}, which synthesizes street-view images from arbitrary camera positions.
Note that on the VIGOR dataset, the positions of street-view images are not always centered in the input satellite patch. Therefore, we limit our comparisons with methods~\cite{Pix2Pix,regmi_cross-view_2019,Shi2022pami}, which can only synthesize single images at the center of the input satellite images, to the CVACT and CVUSA datasets.

\begin{table*}[t!]
\centering
\caption{Quantitative comparison on image level metric on the CVUSA, CVACT, VIGOR, and VIGOR-OOD test set. For Sat2Density and Ours, `*' indicates that for each satellite image input, we randomly select a sky illumination feature from the training set as the illumination input for rendering street views, which provides a fair comparison, and `\dag' means we input ground-truth sky illumination feature, shows the effect of illumination feature on the metric.
}
\resizebox{0.98\linewidth}{!}{
\begin{tabular}{c|c|cc|c|c|c|cc}
\toprule
    \multirow{2}{14mm}{Dataset} & \multirow{2}{*}{Method} & \multicolumn{2}{c|}{Realism Evaluation}  &  Semantic & Structure & Pixel
    & \multicolumn{2}{c}{Perceptual Similarity}  \\
        &     &
         FID$\downarrow$ & KID$\downarrow$ & DINO$\uparrow$  & SSIM$\uparrow$ & PSNR$\uparrow$  & $P_{\text{alex}}\downarrow$ &  $P_{\text{squeeze}}\downarrow$  \\   \midrule

        \parbox[t]{14mm}{\multirow{6}{*}{CVUSA}} 
    & Pix2Pix~\cite{Pix2Pix}  &70.33&.061&0.334 &0.295&13.48&0.509&0.390 \\
     & XFork~\cite{Regmi_2018_CVPR} &63.42&.063 & 0.353 &0.287&13.68&0.501&0.514 \\
     & Shi \etal~\cite{Shi2022pami} &55.37&.059&0.397 &\textbf{0.345}&\textbf{13.75}&0.464&0.351 \\  
     & Sat2Density* &\underline{53.29}&\underline{.047}&\underline{0.401}&\underline{0.330}&\underline{13.45}&\underline{0.457}&\underline{0.346} \\
     & Ours* &\textbf{18.64}& \textbf{.013}&\textbf{0.412} &0.323&13.15&\textbf{0.434}&\textbf{0.344} \\ \cline{2-9} 
     & Ours\textsuperscript{\dag} &19.29&.014 &0.436&0.352&14.84&0.371&0.288 \\ \hline

        \parbox[t]{14mm}{\multirow{6}{*}{CVACT}} 
    & Pix2Pix~\cite{Pix2Pix} &60.33&.055 &0.449 & 0.385 & 14.38  & 0.465 & 0.310  \\
     & XFork~\cite{Regmi_2018_CVPR}  &57.25&.053 &0.468& 0.371 & 14.50  & 0.464 & 0.326  \\
     & Shi \etal~\cite{Shi2022pami}  &43.68&.040 &0.522& 0.427 & \underline{14.59}    & 0.406 & 0.271 \\
     & Sat2Density* &\underline{40.89}&\underline{.034} &\underline{0.530} &\textbf{0.447}&\textbf{14.59}&\underline{0.392}&\textbf{0.258} \\
     & Ours* &\textbf{22.76}&\textbf{.016}&\textbf{0.544}&\underline{0.428}&14.12&\textbf{0.389}&\underline{0.269} \\  \cline{2-9} 
     & Ours\textsuperscript{\dag} &23.20&.016&0.571 &0.458&16.23&0.315&0.214 \\ \hline

 \multirow{3}{14mm}{VIGOR}
      & Sat2Density* &63.53&.049&0.485&0.337&13.11&0.432&0.355 \\
     & Ours* &\textbf{28.08}&\textbf{.020}&\textbf{0.507}&\textbf{0.345}&\textbf{13.14}&\textbf{0.399}&\textbf{0.312} \\  \cline{2-9} 
     
     & Ours\textsuperscript{\dag} &26.11&.019 &0.530&0.391&15.25&0.336&0.256 \\ \hline

 \multirow{3}{14mm}{VIGOR-ood}
      & Sat2Density* &85.66&.079&0.451&0.321&12.48&0.453&0.368 \\
     & Ours* &\textbf{40.85}&\textbf{.035}&\textbf{0.465} &\textbf{0.343}&\textbf{12.51}&\textbf{0.436}&\textbf{0.343} \\   \cline{2-9} 
     & Ours\textsuperscript{\dag} &42.48&.038&0.485 &0.377&14.01&0.385&0.295 \\ \bottomrule
      
\end{tabular}
}

\label{tab:result}
\vspace{-1.5mm}
\end{table*}

\subsubsection{Comparison on Video Results}

We compare our video results with our preliminary model, Sat2Density. Qualitative comparisons are provided in Fig.\ref{fig:demo_vigor} and Fig.~\ref{fig:demo}, and 
quantitative results from user studies are shown in Fig.~\ref{fig:user_study}.

The user study in Fig.~\ref{fig:user_study} highlights a clear preference for \method over Sat2Density across both suburban and urban scenarios. On the CVACT dataset, which features suburban environments with roads, trees, and low-density housing (often partially occluded by vegetation), \method achieves approval ratings of 79\% for Quality, 82\% for Consistency, and 79\% for Faithfulness. On the more complex VIGOR dataset, which contains diverse urban scenes with dense buildings and rich textures, these ratings increase significantly to 95\% for Quality and Consistency, and 92\% for Faithfulness. These results confirm \method’s superior performance, particularly in handling complex scenes.

Qualitative results in Fig.~\ref{fig:demo_vigor} and Fig.~\ref{fig:demo} further support these findings. We showcase selected video frames from both datasets, with full videos available on our project page. On CVACT, both methods capture major features such as greenways, trees, and landmarks. However, Sat2Density often suffers from inconsistent sky appearance and lower frame quality, while \method maintains visual stability and coherence throughout the sequence. Tree structures in particular appear more consistent in \method’s outputs during playback.
On VIGOR, the advantages of \method are even more apparent. It produces higher-quality frames and better temporal consistency, especially in fine-grained structures like crosswalks and buildings. For instance, as shown in the second image set, crosswalks generated by \method are more faithful to the satellite reference. These improvements are credited to our enhanced architecture and training strategy.

To the best of our knowledge, beyond Sat2Density and our proposed \method, only Sat2Vid~\cite{sat2vid} and Sat2Scene~\cite{li2024sat2scene} aim to generate videos from a satellite image. However, both rely on building height maps and focus primarily on rendering facades and road surfaces, limiting scene diversity. Due to the unavailability of their full datasets and code, we are unable to conduct comparisons.

Overall, unlike prior work that focuses narrowly on buildings, our method generates a diverse range of scene elements, including trees, green belts, and landmarks, resulting in videos that are more faithful to satellite inputs and visually realistic. Moreover, \method does not require additional supervision such as height maps or metric depths, making it more versatile and broadly applicable.

\begin{figure*}[h]
\centering
\includegraphics[width=0.88\linewidth]{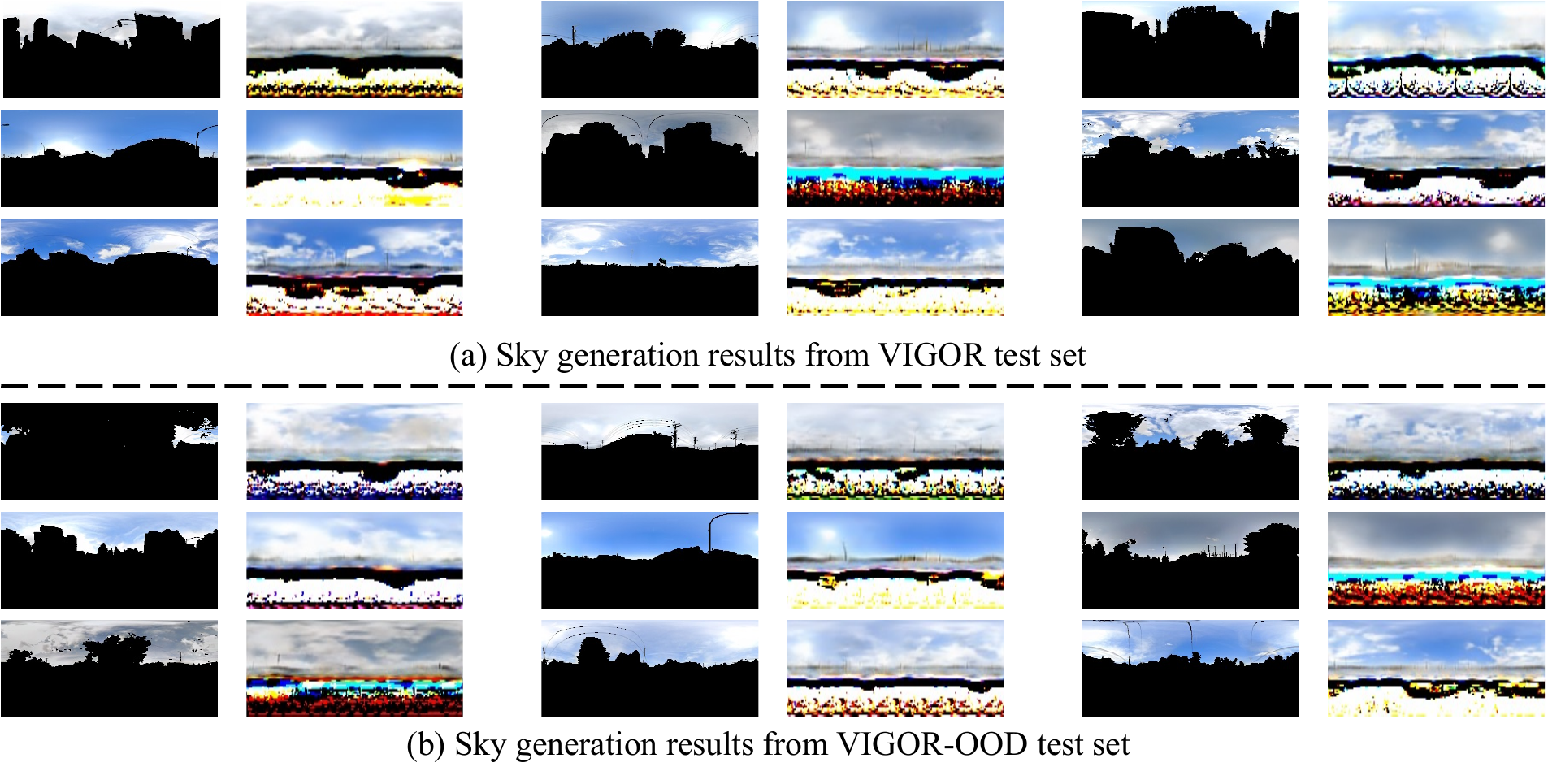}
\caption{ Sky generation from sky-illumination inputs for (a) in-domain and (b) OOD scenarios. In each pair, the left shows a real sky image, while the right presents a sky synthesized from the extracted sky-region histogram. In-domain cases may display higher resemblance due to near-time captures relative to the training data, while OOD cases consistently maintain plausible tone, lighting, and cloud distribution, demonstrating robust generalization of the sky generation.}
\label{Fig: ood_generation}
\end{figure*}

\subsubsection{Comparison on Image Results}

The quantitative image-level results in Tab.~\ref{tab:result} demonstrate the superior performance of our method. Across all four test sets, even when using randomly sampled illumination inputs from the training set (denoted by `*’), our approach consistently outperforms existing methods in perceptual, semantic, and realism metrics. Notably, Sat2Density++ achieves greater gains over Sat2Density on the complex urban test sets (VIGOR-OOD and VIGOR) than on the simpler suburban ones, highlighting its improved capability in modeling complex scenes. These improvements stem from our more effective 3D representation learning design.

While Sat2Density++ shows slightly lower PSNR and SSIM scores than Sat2Density on CVUSA and CVACT, and also underperforms Shi et al.~\cite{Shi2022pami} on CVUSA, this does not indicate true performance drops. Instead, it reflects the limitations of low-level metrics, whose sensitivity to appearance variations in homogeneous regions becomes especially problematic in suburban panoramas.
This is further evidenced by the comparison using real versus random illumination inputs (denoted by `{\dag}’ in Tab.~\ref{tab:result}). With real illumination features, perceptual, PSNR, and SSIM scores improve significantly, while realism metrics such as FID and KID remain stable. This demonstrates the sensitivity of low-level metrics to appearance changes, even when scene semantics are preserved. Realism metrics, based on distributional similarity, offer a more robust measure of perceptual quality. Together, these findings confirm that \method effectively captures and utilizes illumination information, enabling visually consistent and illumination-controllable image synthesis, as illustrated in Fig.~\ref{fig:illumination} and Fig.~\ref{fig:single compare}.

\begin{figure*}[!t]
    \centering
    \includegraphics[width=0.95\textwidth]{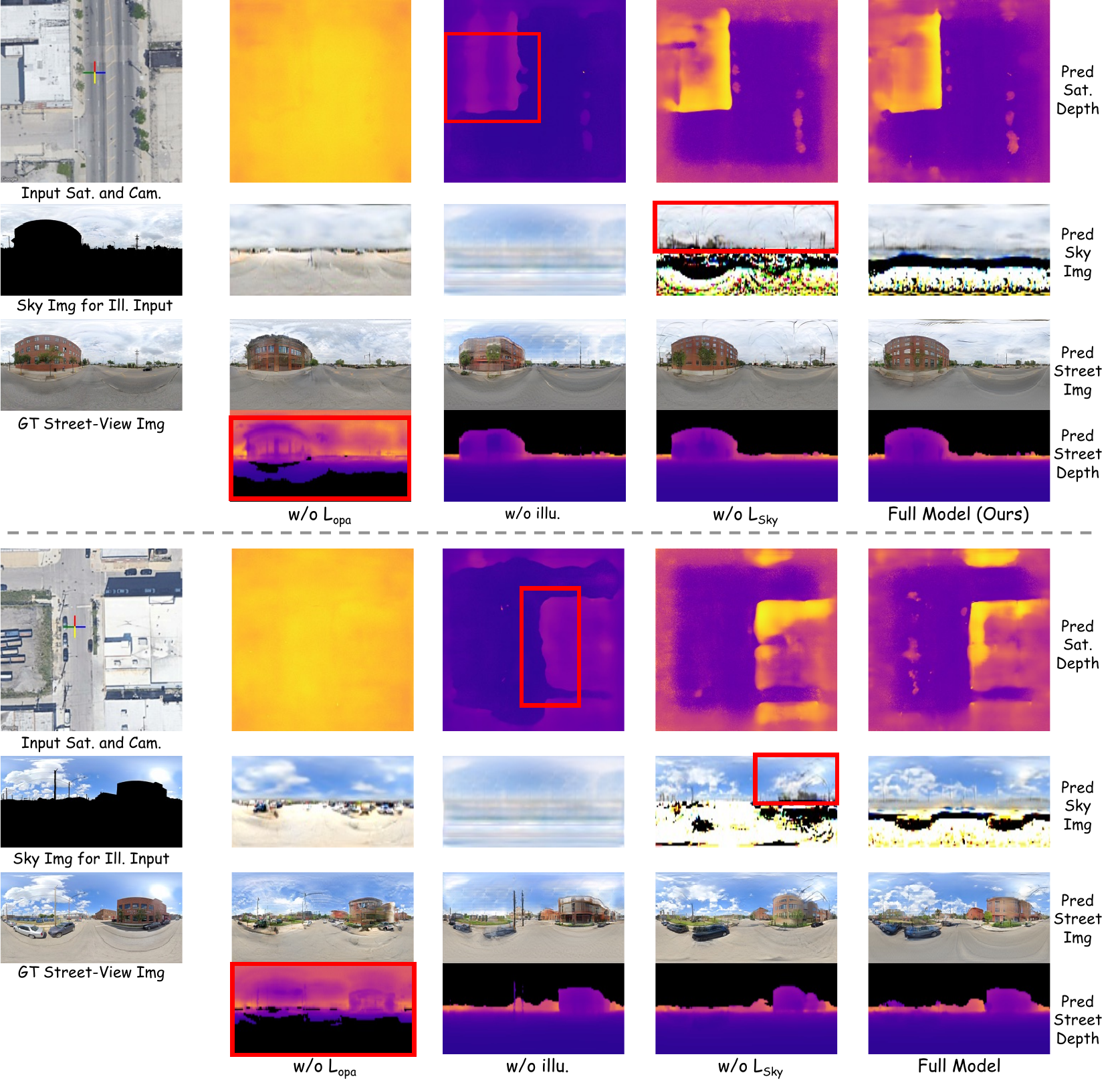}
    \caption{Ablation study for the proposed strategies on the VIGOR dataset. The first column is the input satellite image, the cross point in the satellite image is the input camera position, the masked sky image for sky illumination input,  and the ground truth street-view image. The red boxes show the failure regions.
     {\em The video ablation can be seen on the \href{https://qianmingduowan.github.io/sat2density-pp//}{\texttt{project page}}.}}
    \label{fig:ablation study}
\end{figure*}


\begin{table}[t!]
\centering
\caption{\textbf{Ablation of illumination inputs on VIGOR-OOD test set for satellite- and street-view generation.} Each row sets the illumination code for the two branches (columns ``Satellite'' and ``Street View''): \emph{random} draws a vector $z\!\sim\!\mathcal{N}(0,I)$, \emph{null-style} uses the null-style vector $\mathbf{w}_0$, and \emph{valid} supplies the ground-truth illumination when training. Bold denotes the best configuration (\textbf{Satellite}=\emph{null-style}, \textbf{Street}=\emph{valid}).}
\begin{tabular}{cc|ccc}
\toprule
       Satellite & Street View  & DINO$\uparrow$   & FID$\downarrow$ &KID$\downarrow$   \\ 
      \midrule
    Random & Random &0.439&73.68&.073\\ 
    Valid & Valid &\underline{0.446}&68.03&.065 \\ 
    Null-Style  & Random &0.439&\underline{64.98}&\underline{.060} \\ 
    Null-Style  & Valid   &\textbf{0.465}&\textbf{40.85}&\textbf{.035}  \\ 
    \bottomrule
\end{tabular}
\label{tab:nerf}
\vspace{-1.5mm}
\end{table}


\begin{table}[t!]
\centering
\caption{Ablation results on VIGOR-ood test set.  `w/o $\mathcal{L}_\text{opa}$' means training without non-sky opacity loss, `w/o illu.' means no illumination input $f_\text{ill}$, and replace the  $f_\text{ill}$ and $\mathbf{w}_0$ with a random noise $\mathbf{z}$.
`w/o $\mathcal{L}_\text{sky}$' means removing the sky reconstruction loss.
`w/o sat. view loss' means removing the sat. discriminator and sat. view reconstruction loss.  
`w/o sky branch' means we remove the sky generation branch, use the tri-plane NeRF to render a full street-view image.
`w. vanilla NeRF' means we replace the illumination adaptive tri-plane decoder with a vanilla tri-plane decoder.
}
\begin{tabular}{lccc}
\toprule
       & DINO$\uparrow$ & FID$\downarrow$ &KID$\downarrow$   \\ \midrule
    w/o $\mathcal{L}_\text{opa}$ &0.460&52.30&.048 \\
    w/o illu.  &0.439&73.68&.073  \\
    w/o $\mathcal{L}_\text{sky}$  &0.464&48.78&.043  \\
    w/o $\text{illu.}$ + $\mathcal{L}_\text{opa}$ + $ \mathcal{L}_\text{sky}$  &0.465&59.81&.057 \\ \hline
    w/o sat. view loss &\textbf{0.469}&\textbf{38.71}&\underline{.036} \\\hline
     w/o sky branch &0.457&50.10&.045 \\
     w. vanilla NeRF & 0.460& 48.81& .043 \\ \hline
    Full model  &\underline{0.465}&\underline{40.85}&\textbf{.035}  \\
    \bottomrule
\end{tabular}

\label{tab:ablation study}
\vspace{-1.5mm}
\end{table}

\subsection{Generalizability of sky generation}\label{sec.sky Generalizability}
We included several visualizations of sky generation from sky illumination input on both in-domain and out-of-domain sets, as shown in Fig.~\ref{Fig: ood_generation}. In the VIGOR in-domain test set, some of the training and test data were captured on the same day under nearly identical lighting conditions. Consequently, it is possible to find cases where the generated sky image closely resembles the original sky image corresponding to the sky illumination input, as the training set might have encountered the same sky illumination input. In out-of-domain testing, however, such cases do not occur. The color histograms of sky regions not seen during training can still produce well-rendered sky areas. We observe a high similarity in sky lighting between the generated and real images, demonstrating the robust generalization capability of our sky illumination input.

\subsection{Ablation Study}\label{sec.ablation}
\begin{figure*}[h]
    \centering
    \includegraphics[width=0.9\textwidth]{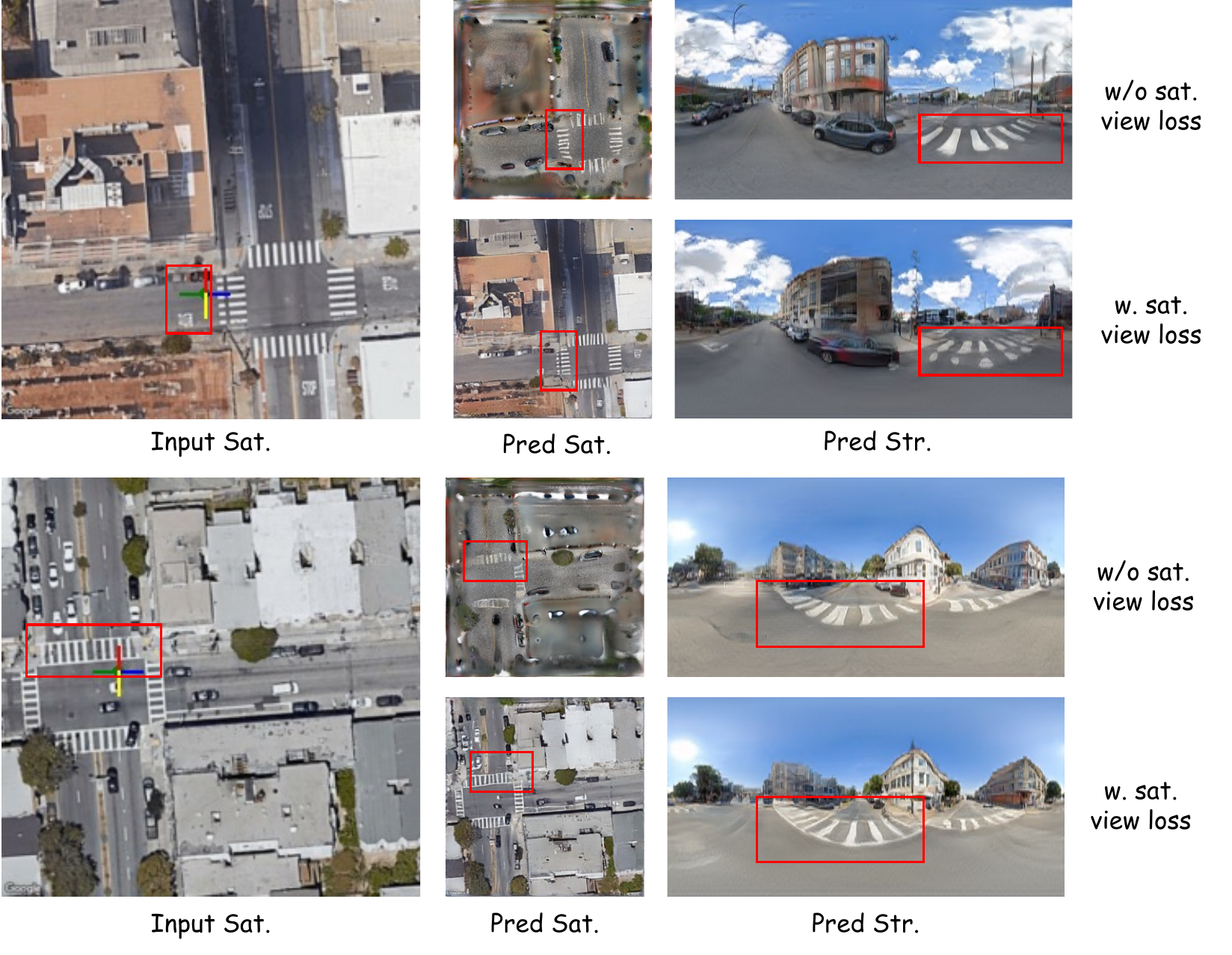}\\
    \caption{
    Ablation study on satellite-view loss.``w/o sat. view loss'' denotes training without  $\mathcal{L}_{\text{sat}}$ and $\mathcal{L}_{D_\text{sat}}$.
    }
    \label{fig:sat_ablation}
\end{figure*}

\begin{figure*}[h]
    \centering
    \includegraphics[width=0.95\textwidth]{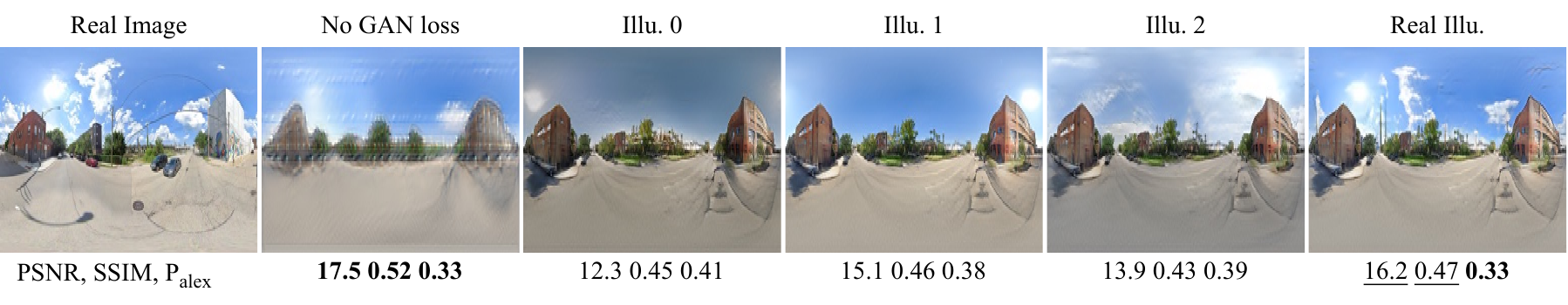}
     \caption{Comparison of image synthesis quality. The second image, trained without GAN loss, attains higher metric scores but yields visibly poorer results. In contrast, the full model shows negligible differences in visual and semantic quality under varying illumination inputs, demonstrating robustness and consistency. The first image is the ground truth reference.}
    \label{fig:single compare}
\end{figure*}

In this section, we will discuss the contributions of key components within \method. The ablation study was conducted on the VIGOR-OOD test set. We present the quantitative ablation in Tab.~\ref{tab:nerf}, Tab.~\ref{tab:ablation study}, Tab.~\ref{tab:ablation GAN} and qualitative ablation in Fig.~\ref{fig:ablation study} and Fig.~\ref{fig:sat_ablation}, Fig.~\ref{fig:single compare}. 
Unless otherwise stated, all quantitative comparisons in the ablation study are conducted under the same setting as '\dag' in Tab.~\ref{tab:result}.

\subsubsection{Sky Illumination Modeling}
The sky illumination input, illumination-adaptive tri-plane decoder, and illumination-adaptive 2D sky generation module work together to mitigate the effects of view-specific lighting in street-view images during training, while also enabling controllable illumination synthesis at inference. Tab.~\ref{tab:nerf} summarizes different design choices for modeling illumination in the satellite and street-view branches.

As shown in Tab.~\ref{tab:nerf}, the poorest performance occurs when both satellite and street views are generated using random noise, completely ignoring the inherent illumination differences between the two views. The corresponding visual results are shown in the “w/o illu.” column of Fig.~\ref{fig:ablation study}. Compared to the full model, the predicted geometry from the satellite view shows larger deviations along object boundaries and more angular artifacts in depth, as highlighted in the red box in the first row. This demonstrates that failing to account for differing imaging conditions leads to degraded 3D representations and lower-quality video generation.
A slight improvement is observed when using the null-style vector for satellite-view generation, while still using random noise for the street view. In this case, the model becomes aware of cross-view appearance differences, but lacks proper compensation through illumination-aware modeling in the street view.

Our full model, which uses a null-style vector for satellite-view generation and the valid sky histogram for the street-view illumination input, achieves the best performance. This configuration fully leverages illumination cues from both views, leading to more accurate geometry, better visual quality, and controllable street-view synthesis.

\begin{table*}[t!]
\centering
\caption{ Ablation results on VIGOR-ood dataset for GAN loss. }
\resizebox{0.75\linewidth}{!}{
\begin{tabular}{lcccccccc}
\hline
      Comparison &PSNR$\uparrow$& SSIM$\uparrow$ &$P_{\text{alex}}\downarrow$ &  $P_{\text{squeeze}}\downarrow$ & DINO$\uparrow$ & FID$\downarrow$ &KID$\downarrow$   \\ \hline
    w/o GAN loss &\textbf{12.96}&\textbf{0.365}&\textbf{0.395}&0.353&0.340&205.62&.239 \\\hline
    Full model  &12.51&0.343&0.436&\textbf{0.343}&\textbf{0.465}&\textbf{40.85}&\textbf{.035}  \\
    \hline
\end{tabular}}
\label{tab:ablation GAN}
\vspace{-1.5mm}
\end{table*}

\subsubsection{Illumination Adaptive Tri-plane NeRF}
We experimented by replacing the illumination-adaptive tri-plane decoder with a vanilla tri-plane decoder. In this case, the sky illumination input, initially intended to affect both sky region and ground appearances under varying daylight, was limited to adjusting only the sky. Using a vanilla tri-plane decoder resulted in a noticeable decline in image quality compared to the illumination-adaptive decoder, as shown in Tab.~\ref{tab:ablation study}. This occurs because NeRF's volumetric rendering technique, which generates pixel values based on light-scene interactions, relies on the consistent appearance of multi-view data~\cite{NERF}. Inconsistent lighting prevents parameter optimization. 

\subsubsection{Dual Branch Design}
\begin{figure}[h]
\centering
\includegraphics[width=0.9\linewidth]{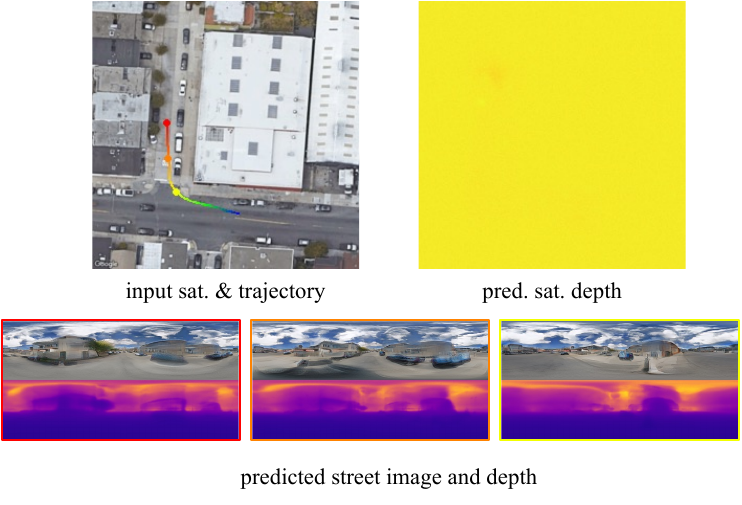}
\caption{
Results without the sky branch: includes predicted satellite-view depth and street-view image with depth.
}
\vspace{-2mm}
\label{Fig: no_sky_branch}
\vspace{-2mm}
\end{figure}

When the sky branch is removed, the tri-plane NeRF must generate both ground and sky regions simultaneously, presenting two major challenges. First, the tri-plane NeRF can only express a limited spatial range~\cite{eg3d2022}, meaning the rendered sky region is given depth and becomes part of the NeRF-generated scene. Consequently, any movement of the camera causes changes in the sky region within the generated panoramic image, which is incorrect because the sky should remain consistent regardless of camera movement, as shown in Fig.~\ref{Fig: no_sky_branch}. Second, it is physically incorrect to represent both the infinitely distant sky and the finite ground regions within a limited 3D space. This increases the model's learning burden and degrades learning quality, significantly reducing the visual realism of the generated images and videos, as quantitatively demonstrated in Tab.~\ref{tab:ablation study}.

\begin{table}[h]
\centering
\vspace{-2mm}
\caption{Ablation results for input satellite-view reconstruction on VIGOR-OOD test set. } 
\vspace{-2mm}
\begin{tabular}{lcc}
\toprule
       & PSNR$\uparrow$ & SSIM$\uparrow$ \\ \hline
     w/o sat. view loss &27.5 &0.21 \\ \hline
    Full model  & \textbf{30.0} &\textbf{0.56}  \\
    \bottomrule
\end{tabular}

\label{tab:ablation study for Sat. view loss satview}
\vspace{-1.5mm}
\end{table}

\begin{figure*}[h]
\centering
\includegraphics[width=0.9\linewidth]{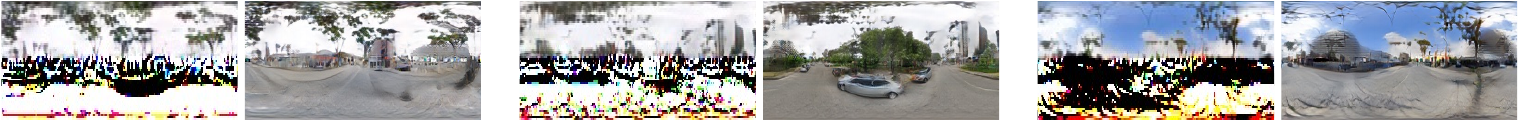}
\caption{
Visualization of results without the sky-region reconstruction loss. We presented three sets of images; in each set, the left image is the generated sky image, and the right is the final generated street-view image.
}
\label{Fig: no_sky_loss}
\end{figure*}
\subsubsection{Loss Functions}

\noindent\textbf{Role of $\mathcal{L}_{\text{opa}}$.} The non-sky opacity loss function $\mathcal{L}_{\text{opa}}$ assists in delineating the learning of the 3D ground scene and the 2D sky region within the density field. By employing $\mathcal{L}_{\text{opa}}$, the model is enabled to approximate non-sky areas with opacity values near 1, while sky regions tend towards an opacity of 0. Omitting $\mathcal{L}_{\text{opa}}$ leads to significant generation errors: the 2D sky generator erroneously renders ground features, the satellite and street-view depths rendered are inaccurate compared to the full model, the road region remains static in the generated video, and there is misalignment between the sky and the upper half of the image, as illustrated in the second column of Fig.~\ref{fig:ablation study}. Without this loss function, the model struggles to differentiate the generation roles of the tri-plane and sky generators under only image-level supervision, resulting in incorrect region generation.

\noindent\textbf{Role of $\mathcal{L}_{\text{sky}}$.} The sky loss $\mathcal{L}_{\text{sky}}$ is crucial for 2D sky image generation. The incorporation of $\mathcal{L}_{\text{sky}}$ leads to notable improvements in realism metrics such as FID and KID. As shown in the  Fig.~\ref{Fig: no_sky_loss} and fourth column of Fig.~\ref{fig:ablation study}, 
without the proper guidance of sky loss, ground content that should be rendered by the tri-plane NeRF may incorrectly appear in the sky images, such as trees and buildings. This causes the model to fail in accurately distinguishing the generated content between the two branches, leading to unrealistic outcomes.

\noindent\textbf{Role of $\text{illu.}$ + $\mathcal{L}_\text{opa}$ + $\mathcal{L}_\text{sky}$}. In \method, we employ illumination modeling, which includes illumination input, the illumination-adaptive NeRF, and the illumination-adaptive 2D sky generator module, to mitigate the effects of illumination on 3D representation learning. Additionally, we utilize $\mathcal{L}_\text{opa}$ and $\mathcal{L}_{\text{sky}}$ to segregate the optimization process of 3D scenes from 2D sky regions during training. When these components are removed, the model can be perceived as \textit{a vanilla image-conditioned 3D-aware generation model}. As shown in Tab.~\ref{tab:ablation study}, performance scores significantly decrease in comparison to the full model, and the generated street-view videos are inconsistent and of poor quality, as demonstrated on our project page. This underscores that these training strategies and module designs, specifically tailored to address the SatStreet-view synthesis task's challenges, are effective in introducing an image-conditioned 3D-aware generation model.


\noindent\textbf{Role of satellite-view loss.}
We apply satellite-view reconstruction and GAN losses to align the generated 3D appearance with the input satellite image \( I_{sat} \). These losses ensure that the images rendered from the learned radiance field at the satellite viewpoint closely match \( I_{sat} \), thereby guaranteeing that the output street-view video remains faithful to the satellite image.
As illustrated in Tab.~\ref{tab:ablation study}, omitting the satellite-view reconstruction loss results in slightly improved FID and DINO scores. However, without photometric supervision from the input satellite perspective, the learned 3D representation is prone to deviating from the information provided by the satellite view. 
Quantitatively, Tab.~\ref{tab:ablation study for Sat. view loss satview} shows that removing the satellite-view loss leads to a notable drop in satellite image reconstruction quality on the VIGOR-OOD test set (PSNR decreases from 30.0 to 27.5, and SSIM from 0.56 to 0.21).
In Fig.~\ref{fig:sat_ablation}, the predicted satellite images show deterioration in the absence of satellite supervision, lacking consistency with the input satellite imagery, such as in crosswalks, trees,
and building appearances, and the generated zebra crossings in the predicted street-view images exhibit a different style compared to those in the input satellite image. This indicates that incorporating the satellite-view losses significantly enhances the faithfulness of the generated ground images and videos to the input satellite image.

\noindent\textbf{Role of GAN loss.}
We use GAN loss~\cite{GAN} to ensure that the rendered images appear realistic from any viewpoint. As depicted in Tab.~\ref{tab:ablation GAN}, omitting the GAN loss results in higher scores on metrics such as PSNR, SSIM, and $P_{\text{alex}}$. However, the realism of the generated images is compromised, as indicated by poorer scores on realism metrics such as FID and KID. As shown in Fig.~\ref{fig:single compare}, the generated images appear very blurry. This underscores the importance of GAN loss in enhancing the authenticity of the generated images and videos.


\begin{figure*}[ht]
    \centering
    \includegraphics[width=0.95\textwidth]{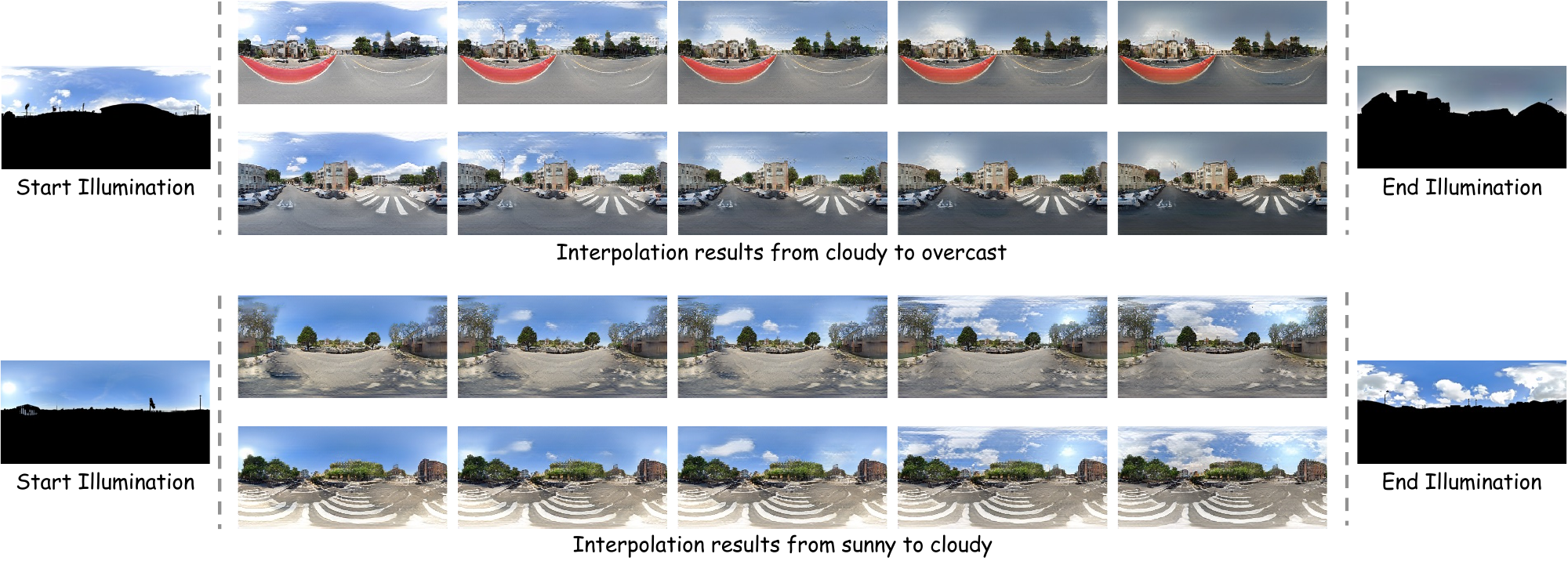}\\
    \vspace{-1mm}
    \caption{ Illumination interpolation street-view image synthesis.}
    \label{fig:illumination}
    \vspace{-1em}
\end{figure*}

\begin{figure*}[ht]
    \centering
\includegraphics[height = 0.15\linewidth]{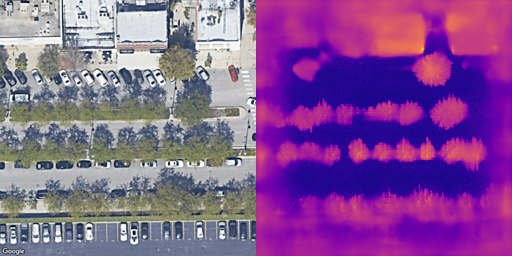}
\includegraphics[height = 0.15\linewidth]{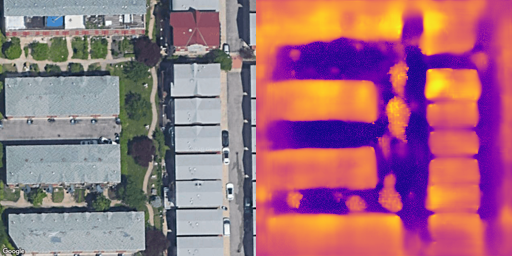}
\includegraphics[height = 0.15\linewidth]{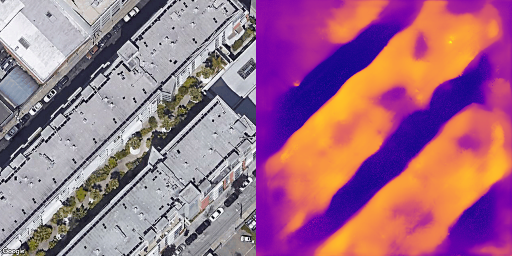}\\
\vspace{0.2em}
\includegraphics[height = 0.15\linewidth]{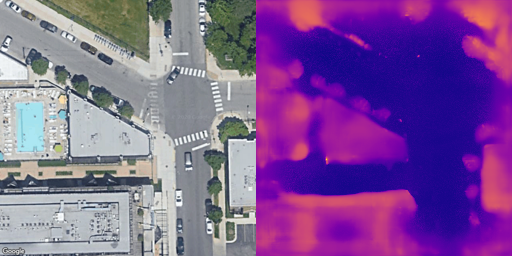}
\includegraphics[height = 0.15\linewidth]{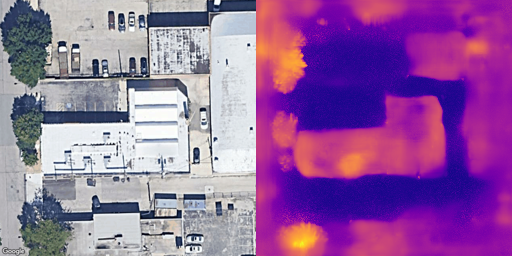}
\includegraphics[height = 0.15\linewidth]{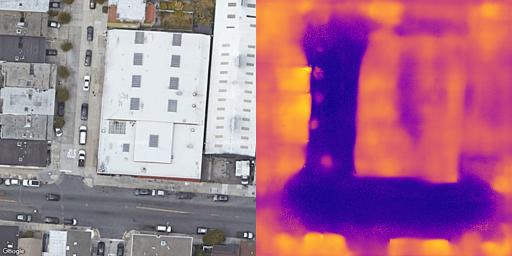}\\
\vspace{0.2em}
\includegraphics[height = 0.15\linewidth]{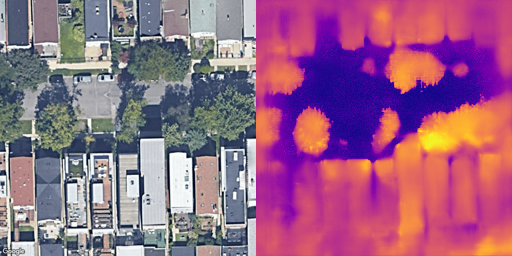}
\includegraphics[height = 0.15\linewidth]{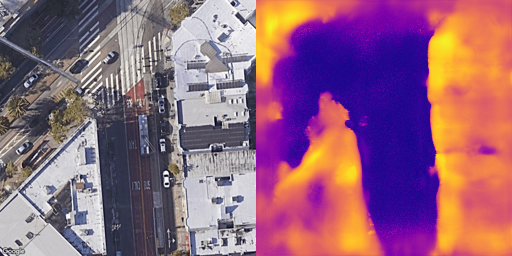}
\includegraphics[height = 0.15\linewidth]{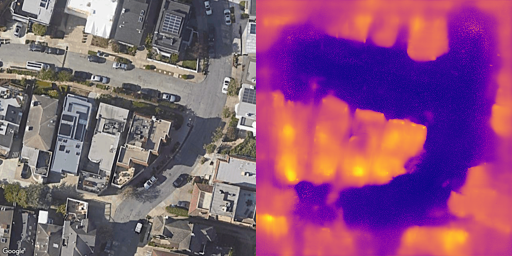}
    \caption{Applications on satellite depth prediction. In the predicted depth image, colors closer to gold represent higher elevations, while colors closer to purple correspond to lower heights.
}
    \label{fig: Depth Prediction}
    \vspace{-1em}
\end{figure*}

\subsection{Applications}\label{sec.application}
To demonstrate the practical utility and robustness of our proposed model, we explore downstream applications. These applications serve to validate the effectiveness of our design and highlight the potential for real-world impact.

\noindent\textbf{Illumination-controllable street-view image synthesis.}
One notable application of our model is illumination-controllable street-view image synthesis. As shown in Fig.~\ref{fig:illumination} and Fig.~\ref{fig:single compare}, by manipulating the illumination feature, our model enables precise control over the lighting conditions in the synthesized image while preserving the underlying scene semantics.  For instance, the sky and color attributes of roads, buildings, and foliage can be varied through different illumination settings while their shapes are preserved unchanged. This capability has potential applications in domains such as virtual reality, gaming, and film production, where fine-grained control over scene lighting is desirable.


\noindent\textbf{Mono-depth estimation from satellite imagery.}
Fig.~\ref{fig: Depth Prediction} showcases the ability to predict depth from satellite imagery without explicit metric or relative depth supervision of \method. The inferred depth maps reflect the deep understanding of scene structure,  building structure, and shape of the trees, \etc This ability to capture the fine-grained details of various landscape elements attests to the potential utility in fields such as urban planning and geographical information systems, where depth information gleaned from aerial imagery is invaluable.

\begin{figure*}[t!]
    \centering
    \includegraphics[width=0.95\textwidth]{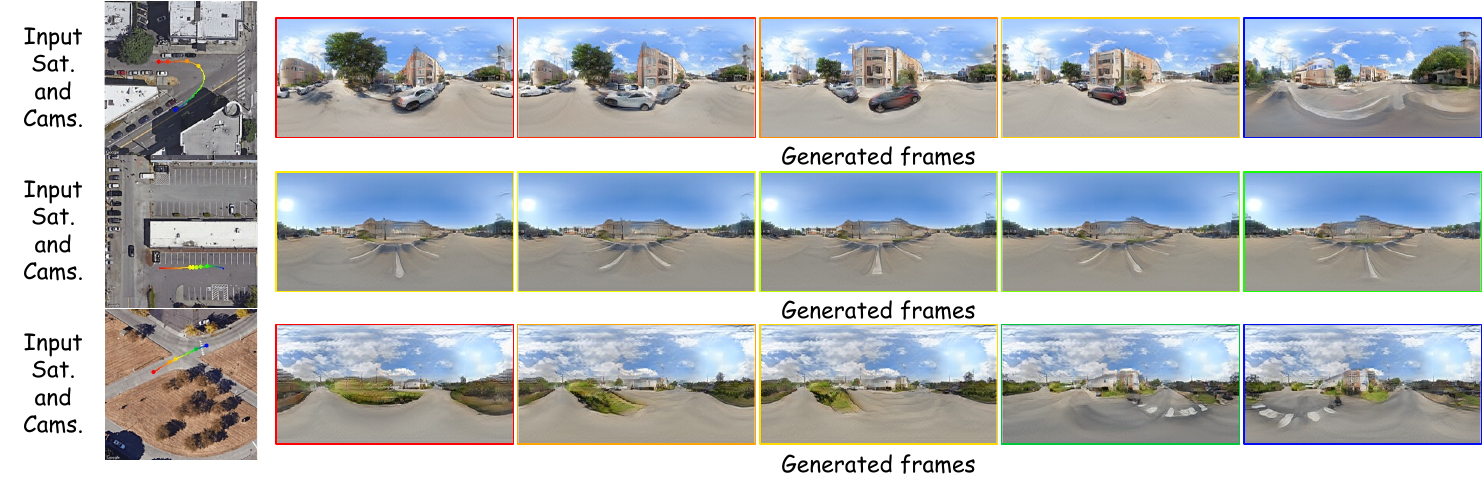}
    \vspace{-1mm}
\caption{Selected frames from the synthesized video in Seattle City. All the trajectories were filmed sequentially from the red to the blue point. {\em The full videos can be seen on the \href{https://qianmingduowan.github.io/sat2density-pp//}{\texttt{project page}}.}}
\label{fig: Seattle}
\vspace{-4mm}
\end{figure*}

\noindent\textbf{Out-domain generalization in Seattle City.}
To assess the generalization capability of our model, we perform a zero-shot evaluation on satellite imagery from Seattle, a city not included in the training data. As shown in Fig.~\ref{fig: Seattle}, our model maintains multi-view consistency and accurately reproduces the layout of the city, including details such as parking spaces and building architecture. This zero-shot generalization ability highlights the potential for our model to be applied to diverse geographic locations without the need for additional training data, which is particularly valuable for tasks such as autonomous navigation.

These downstream applications demonstrate the versatility and effectiveness of our proposed model, showcasing its potential for real-world impact across various domains. By developing a robust and generalizable approach to SatStreet-view synthesis, our work opens up new avenues for leveraging remote sensing data in a wide range of computer vision and remote sensing tasks.


\section{Discussion \& Limitations}

In this section, we explore the challenges faced in our SatStreet-view synthesis approach and discuss the limitations of our current algorithm and data processing pipeline. By highlighting these issues, we aim to provide insights for future research directions.

\textbf{Data challenge.}
Translating satellite imagery to street views within our conditional 3D generation framework necessitates well-aligned imagery, similar to datasets such as CelebAMask-HQ, AFHQ-cat, and Shapenet-car~\cite{CelebAMask-HQ, AFHQ-cat, Shapnet-car}. However, satellite images are not always captured from a nadir perspective, and precise pose calibration of street-view images remains challenging. Moreover, the altitude of ground images can vary considerably, for example, when captured from elevated positions such as overpasses, further complicating their placement in both world and camera coordinate systems, a step that is critical for accurately defining the generative space.

\textbf{Use sky region color histogram to represent the sky illumination.} We acknowledge the potential of enriching illumination inputs with data such as sun position, cloud distribution, or atmospheric radiation to better handle unknown conditions and improve video quality. However, this would deviate from our focus and introduce substantial complexity. Daylight estimation from outdoor imagery is a challenging and ill-posed problem~\cite{lalonde2014lighting,zhang2019all,hold2019deep,DBLP:journals/ijcv/LalondeEN12}, and relying on existing models may add noise while requiring extra effort for our synthesis framework to interpret. Given the demonstrated effectiveness of the simple sky-region histogram, we defer incorporating richer illumination cues to future work.

\textbf{3D space for generation.} We approximate the sky region as the only area in street-view images that extends beyond the satellite scene. Although this approach has proven effective, it lacks precision: areas extending beyond the satellite coverage might also include distant mountains, buildings, and other elements. Accurately distinguishing and modeling these regions is a direction worthy of future exploration, as it would enhance the accuracy of the learned 3D representations.

\section{Conclusion}
In this paper, we studied the task of SatStreet-view synthesis and presented \method, a novel approach to synthesize street-view panorama video from a given satellite image as input. Our \method is built upon a deep understanding of SatStreet-view synthesis, devising an effective learning solution by modeling sky regions and street-view illuminations. It approaches the challenging two-view neural field learning problem with a feedforward 3D-aware generation framework for photorealistic street-view panorama synthesis.
Extensive experiments on the commonly used suburban scene dataset and the recently proposed urban scene dataset VIGOR demonstrated the effectiveness of our proposed \method. To the best of our knowledge, our model is the first that can synthesize multi-view consistent street-view videos from input satellite images, trained without relying on 3D annotations. We hope our proposed method will facilitate 3D representation learning and video generation in outdoor scenes.


{
\bibliographystyle{IEEEtran}
\bibliography{egbib}
}

\end{document}